\documentclass[final, 3p, times, review]{elsarticle}
\usepackage[utf8]{inputenc}

\usepackage{geometry} 
\usepackage{amssymb}
\usepackage{amsmath}
\usepackage{amsthm}

\usepackage{graphicx}
\usepackage{color}
\usepackage{hyperref}
\usepackage[capitalize]{cleveref} 
\usepackage{diagbox}

\usepackage{algorithm}
\usepackage{algpseudocode}
\usepackage{multirow}
\usepackage{booktabs}
\usepackage{threeparttable}
\usepackage{makecell}

\newtheorem{theorem}{Theorem}[section]
\newtheorem{corollary}{Corollary}[section]

\newtheorem{remark}{Remark}[section]

\newcommand{\norm}[1]{\left\lVert#1\right\rVert}
\newcommand{\snorm}[1]{\lVert#1\rVert}
\newcommand{\brac}[1]{\left(#1\right)}
\newcommand{\R}{\mathbb{R}}

\newcommand{\N}{\mathcal{N}}

\newcommand{\0}{\mathbf{0}}
\newcommand{\1}{\mathbf{I}}

\renewcommand*{\d}{\mathop{}\!\mathrm{d}}

\newcommand{\vxi}{{\xi}}
\newcommand{\E}[1]{\mathop{}\!\mathrm{E} [#1]}

\newcommand{\nmu}{\widehat{\mu}}
\newcommand{\nSigma}{\widehat{\Sigma}}
\newcommand{\Nmu}{\widetilde{\mu}}
\newcommand{\NSigma}{\widetilde{\Sigma}}

\usepackage[normalem]{ulem}

\journal{Journal of Computational physics}

\begin{document}

\begin{frontmatter}
\title{DynGMA: a robust approach for learning stochastic \\differential equations from data}
\author[1]{Aiqing Zhu}
\author[1,2]{Qianxiao Li\corref{cor}}
\date{November 2021}
\address[1]{Department of Mathematics, National University of Singapore, 10 Lower Kent Ridge Road, 119076, Singapore}
\address[2]{Institute for Functional Intelligent Materials, National University of Singapore, 4 Science Drive 2, 117544, Singapore}
\cortext[cor]{Corresponding author.
E-mail addresses: zaq@nus.edu.sg, qianxiao@nus.edu.sg}
\date{}

\begin{abstract}
Learning unknown stochastic differential equations (SDEs) from observed data is a significant and challenging task with applications in various fields. Current approaches often use neural networks to represent drift and diffusion functions, and construct likelihood-based loss by approximating the transition density to train these networks. However, these methods often rely on one-step stochastic numerical schemes, necessitating data with sufficiently high time resolution.
In this paper, we introduce novel approximations to the transition density of the parameterized SDE: a Gaussian density approximation inspired by the random perturbation theory of dynamical systems, and its extension, the dynamical Gaussian mixture approximation (DynGMA). 
Benefiting from the robust density approximation, our method exhibits superior accuracy compared to baseline methods in learning the fully unknown drift and diffusion functions and computing the invariant distribution from trajectory data.
And it is capable of handling trajectory data with low time resolution and variable, even uncontrollable, time step sizes, such as data generated from Gillespie's stochastic simulations.
We then conduct several experiments across various scenarios to verify the advantages and robustness of the proposed method.
\end{abstract}
\end{frontmatter}

\section{Introduction}
Dynamical systems with random perturbations can be described by Stochastic Differential Equations (SDE) driven by Brownian noise. This modeling approach serves as a scientific modeling tool applied across a diverse array of disciplines, such as biology, physics, and chemistry. In practical scenarios, the underlying dynamics, including the drift and diffusion functions, are often unknown or only partially understood, while a finite set of discrete observational data is available. For example, various physical experimental systems generate data pertaining to motion tracking \cite{chen2023constructing, chen2022automated, schmidt2009distilling}, microscopic fine-gained physics simulations yield macroscopic coarse-grained statistics \cite{chen2023constructing, dietrich2023learning} and the concentrations of various individuals in an epidemic process can be obtained through Monte Carlo simulations \cite{dietrich2023learning, makeev2002coarse}. In these cases, researchers have to discover interpretable SDE models from data and then infer information such as future behavior, invariant distribution, and energy landscape of the system.

The discovery of dynamical systems, including systems with randomness, from observed data is a well-established task \cite{bongard2007automated, brunton2017chaos, rudy2017data, schmidt2009distilling}. 
SDEs are characterized by their transition densities, which are typically challenging to express in a closed form due to their involvement in solving the Fokker-Planck-Kolmogorov (FPK) equation. As the transition density is essential for likelihood-based parametric estimation of SDEs sampled at discrete times, various approximations have been proposed, e.g., sampling-based approaches \cite{brandt2002simulated, pedersen1995new}, methods based on approximate solutions to the FPK equation \cite{jensen2002transition}, higher-order Ito-Taylor expansion method \cite{kessler1997estimation}, and Hermite polynomials expansion method \cite{ait2002maximum}. Details of the methods on this subject can be found in monographs \cite{iacus2008simulation, prakasa1999statistical}. 
Recently, thanks to the exponential growth of available data and computing resources, machine learning has emerged as a flexible and data-driven approach for learning fully unknown stochastic dynamics \cite{chen2021solving,chen2023constructing,chen2023learning, dietrich2023learning, gu2023stationary,look2022deterministic, solin2021scalable,xu2023modeling}. In addition, SDEs are also employed as a black box model in a diverse range of machine learning tasks \cite{li2020scalable, look2019differential,song2021score, tzen2019neural}.

It has been shown that sampling-based approaches converge to the true transition density as the number of samples increases \cite{brandt2002simulated, pedersen1995new}. However, such approaches are typically used for small-scale parameter scenarios where the dynamics is not fully unknown, and require drawing a large number of sample trajectories to accurately represent the underlying distribution as discussed and numerically validated in \cite{look2022deterministic}. These issues are addressed by developing deterministic Gaussian approximations to the transition density \cite{look2022deterministic, solin2021scalable}. Specifically, Look et al. \cite{look2022deterministic} propose bidimensional moment matching (BMM) to approximate the transition density by a Gaussian density, relying on the layered structure of neural networks. However, the dependence on the layered structure limits its flexibility and makes it difficult to encode partially known physical terms. In addition, Solin et al. \cite{solin2021scalable} propose FPK moment matching (FPK-MM) by Gaussian assumed density approximations of the FPK equation following classical SDE techniques \cite{kushner1967approximations, sarkka2013gaussian}. However, both approaches treat SDEs as a black box and have not thoroughly investigated its performance in learning interpretable governing functions of SDE models. 
Moreover, the Gaussian approximation and the exact density function evidently do not coincide, and the error quantification of existing Gaussian-based approaches remains open.

The most straightforward and efficient Gaussian approximation can be obtained by employing the one-step Euler scheme to the Gaussian approximation of the FPK equation. Alternatively, it can also be obtained through the one-step Euler-Maruyama scheme \cite{dietrich2023learning}. We refer to this approach as the Euler-Maruyama approach in this paper. From this point of view of stochastic numerical integrators, approximations of the transition density beyond Gaussian are derived for other one-step numerical schemes, including Milstein and Runge-Kutta type integrators \cite{dietrich2023learning}. The error of such approximations can thus be deduced from error analysis of the underlying numerical schemes, and numerical results demonstrate the effectiveness of learning SDEs using the proposed approximations \cite{chen2023constructing, dietrich2023learning, lin2023computing}. In addition, Gu et al. \cite{gu2023stationary} derive the loss function directly rather than using likelihood loss, also based on the Euler-Maruyama scheme.
However, methods in this direction may have the limitation of requiring data with sufficiently high time resolution since they are all based on one-step numerical schemes, which will be demonstrated in our numerical experiments later.

\begin{table}[t]
  \centering
    \caption{Comparison between Sampling-based method \cite{brandt2002simulated, pedersen1995new}, BMM \cite{look2022deterministic}, FPK-MM \cite{solin2021scalable}, Integrators-based method \cite{dietrich2023learning} and DynGMA.
    The symbol $\checkmark$ indicates that the method maintains reliability and good performance under these scenarios, or possesses these properties.}
    \resizebox{\linewidth}{!}{
\begin{tabular}{ c|c c c c c c}
  \toprule
  \diagbox{Methods}{Scenarios} &  \makecell[c]{Deterministic\\ approximation}&\makecell[c]{Fully unknown \\ dynamics } & \makecell[c]{Partiall known\\ physical terms  }& \makecell[c]{Interpretable \\ governing function} &\makecell[c]{Low time \\resolution }& \makecell[c]{Theoretically \\ valid}\\
  \hline
  Sampling-based \cite{brandt2002simulated, pedersen1995new} &   & &partially &\checkmark &\checkmark &\checkmark\\
  \hline
  
  BMM \cite{look2022deterministic} & \checkmark & \checkmark & & &  & \\
  \hline

  FPK-MM \cite{kushner1967approximations, sarkka2013gaussian, solin2021scalable} & \checkmark & \checkmark&partially &  &  & \\
  \hline
  Integrators-based \cite{dietrich2023learning} & \checkmark & \checkmark& \checkmark &\checkmark & &\checkmark\\

  \hline
  DynGMA (ours) & \checkmark & \checkmark & \checkmark &\checkmark &\checkmark&\checkmark\\
  \bottomrule
\end{tabular}
}\label{tab:com}
\end{table}

In this paper, we employ neural networks to parameterize the drift and diffusion functions of the fully unknown SDE, and use likelihood-based loss to train the network.
By revisiting the random perturbation theory of dynamical systems \cite{blagoveshchenskii1962diffusion, blagoveshchenskii1961certain}, we propose a novel Gaussian density approximation, and its extension, named as dynamical Gaussian mixture approximation (DynGMA), to approximate the transition densities of the parameterized SDE. 
Specifically, we discretize the time step into several sub-intervals, and employ the proposed Gaussian approximation in each sub-interval. 
We quantify the error of the proposed Gaussian approximation, indicating the advantages of our method compared to the Gaussian approximation obtained via the Euler-Maruyama scheme. 
With DynGMA, we can obtain the automatically-differentiable approximation of the fully unknown drift and diffusion functions of the SDE, and we can then compute the invariant distribution of unknown stochastic dynamics from trajectory data that does not necessarily follow the equilibrium distribution. 
Moreover, our method operates independently of specific structures imposed on stochastic dynamics and thus naturally accommodates the encoding of partially known physical terms, such as generalized Onsager principle \cite{chen2023constructing, yu2021onsagernet}, symplectic structure \cite{greydanus2019hamiltonian}, and GENERIC formalism \cite{zhang2022gfinns}. 
In addition to possessing these advantages, the notable benefit of our approach is its ability to effectively learn SDEs even when using trajectory data with low time resolution. This property enables DynGMA to incorporate multi-step training \cite{williams1989learning} and be extended to a recurrent version without any modification, allowing it to effectively handle data with measurement noise. 
A detailed comparison between our method and the aforementioned approaches is presented in the \cref{tab:com}.

This paper is organized as follows: In \cref{sec:learning sde}, we briefly present the background and the settings of learning SDEs combining neural networks and maximum likelihood estimation. In \cref{sec: Numeriacl methods}, we introduce the proposed DynGMA and the machine learning algorithm in detail. In \cref{sec:Numerical experiments}, we provide several numerical results to investigate the performance of the proposed methods. Finally, we summarize our work in \cref{sec:Conclusions}.

\section{Learning SDEs from data and its applications}\label{sec:learning sde}
In this paper, our focus is on the following non-linear time-invariant stochastic differential equation (SDE) that describes the dynamics of a $D$-dimensional stochastic process $y(t)$,
\begin{equation}\label{eq:sde}
\d y(t) = f(y(t)) \d t + \sigma (y(t)) \d \omega(t).
\end{equation} 
Here, $f: \R^D \rightarrow \R^D$ represents the drift function, $\sigma: \R^D \rightarrow \R^{D \times D}$ is the diffusion function, and $\omega(t)$ is a $D$-dimensional standard Brownian motion inducing randomness. It is assumed that $f$ and $\sigma$ obey Lipschitz-continuity, i.e., $\forall y, \hat{y},\ \norm{f(y) - f(y')} + \norm{\sigma(y) - \sigma(\hat{y})} \leq c_1 \norm{y -  \hat{y}}$ for some constant $c_1$, and linear growth, i.e., $\forall y,\ \norm{f(y)} + \norm{\sigma(y)} \leq c_2(1+ \norm{y})$ for some constant $c_2$, to ensure the existence of the solution of SDE (\ref{eq:sde}); and that the diffusion matrix $\sigma \sigma^{\top}$ is strictly positive definite to ensure the feasibility of the following parametric estimation. 

We consider the scenario where the drift and diffusion functions are fully unknown, but we have access to discrete observations of certain trajectories of $y(t)$. More generally, the given training data is denoted as 
\begin{equation}\label{eq:data}
\{y^n_0, y^n_1, \cdots, y^n_M,\ t_0, t_1, \cdots, t_M \}_{n=0}^N,  
\end{equation}
where $y^n_0 = y^n(t_0), n=1,\cdots, N$ are initial points scattered in the phase space of (\ref{eq:sde}) and the corresponding $y^n_m = y^{n}(t_m), m=1,\cdots, M$ is the evolution of (\ref{eq:sde}) at time $t_m$ starting at $y^n_0$, namely,
\begin{equation*}
y^n(t) = y^n_0 + \int_{t_0}^t f(y^n(\tau)) \d \tau + \sigma (y^n(\tau)) \d \omega(\tau).
\end{equation*}
The purpose of this paper is to develop a machine learning method for identifying the drift function $f$ and the diffusivity function $\sigma$, as well as for subsequent tasks of time series prediction and invariant distribution computation.

Representing the unknown governing functions by neural networks $f_{\theta}$ and $\sigma_{\theta}$ with parameter $\theta$, we obtain 
\begin{equation}\label{eq:sdenet}
\d x(t) = f_{\theta}(x(t)) \d t + \sigma_{\theta}(x(t)) d \omega(t),
\end{equation}
of which solution $x(t)$ is a random variable and follows the transition density $p(t, x(t)|0,  x(0); \theta)$. Given the training data (\ref{eq:data}) and the transition density of parameterized SDE (\ref{eq:sdenet}), we can employ maximum likelihood estimation (MLE) to infer the parameters $\theta$:
\begin{equation}\label{eq:mle}
\begin{aligned}
\theta^{*}  &= \arg \max_{\theta}  \sum_{n=1}^N\sum_{m=1}^M \log p(\Delta t_m, y^n_m|0,  y^n_{m-1}; \theta),
\end{aligned}
\end{equation}
where $\Delta t_m = t_m - t_{m-1}$. We typically require an approximation of the transition density, denoted as $p_h(\Delta t, x|0, x_{0})$. For example, according to the Euler-Maruyama scheme, we can obtain the most straightforward and efficient Gaussian approximation \cite{dietrich2023learning}:
\begin{equation}\label{eq:GaussApp}
p_h(\Delta t, x|0, x_{0}) = \N(x|x_0 + \Delta t f_{\theta}(x_0), \Delta t\sigma_{\theta}(x_0) \sigma^{\top}_{\theta}(x_0)).
\end{equation}
Then the corresponding MLE objective is rewritten as 
\begin{equation}\label{loss}
\begin{aligned}
\theta^{*}  &=\arg \max_{\theta} \sum_{n=1}^N\sum_{m=1}^M \log p_{h}(\Delta t_m, y^n_m|0,  y^n_{m-1}; \theta).
\end{aligned}
\end{equation}
In addition, if an approximation of the transition density is also valid for multiple steps, we instead introduce the multi-step training, a well-known technique in time series analysis \cite{williams1989learning}, by maximizing the likelihood objective:
\begin{equation}\label{multi-steploss}
\theta^{*} = \arg \max_{\theta}\sum_{\substack{\gamma \in \Gamma\\ \Gamma \subset \{1, \cdots, M\}}}\sum_{n=1}^N\sum_{m=1}^{\lfloor M/\gamma\rfloor} \log p_{h}(\Delta t_{m,\gamma}, y^{n}_{m\gamma}|0,  y^{n}_{(m-1)\gamma}; \theta),
\end{equation}
where $\Delta t_{m,\gamma} = t_{m\gamma} - t_{(m-1)\gamma}$.

After optimization, the returned networks $f_{\theta^*}$ and $\sigma_{\theta^*}$ serve as an automatically-differentiable approximation of the target drift $f$ and diffusion $\sigma$, respectively.  Subsequently, we can consider further applications, such as predicting the dynamics by solving the learned SDE and computing the invariant distributions that solve the stationary FPK equation:
\begin{equation*}
- \sum_{d=1}^D \frac{\partial }{\partial x_d} \{ [f_{\theta^*}(x)]_d  p(x) \} + \sum_{d=1}^D \sum_{d'=1}^D \frac{\partial^2 }{\partial x_{d_1}\partial x_{d'}} \{ [\sigma_{\theta^*}(x)\sigma_{\theta^*}^{\top}(x)]_{dd'}  p(x) \} = 0,\ \int_{\R^D} p(x) \d x=1.
\end{equation*}
As the transition density is essential for the MLE and existing methods mostly rely on Gaussian approximations \cite{dietrich2023learning,look2022deterministic, solin2021scalable} or one-step numerical schemes \cite{dietrich2023learning}, we will quantify the errors associated with Gaussian approximations based on the random perturbation theory of dynamical systems \cite{blagoveshchenskii1962diffusion,blagoveshchenskii1961certain}, and develop novel approximations that allow moderately large time steps in the next section.

\section{Numerical methods}\label{sec: Numeriacl methods}
\subsection{Dynamical Gaussian mixture approximation}\label{DynGMA}
\begin{figure}[!ht]
\centerline{\includegraphics[width=1\linewidth]{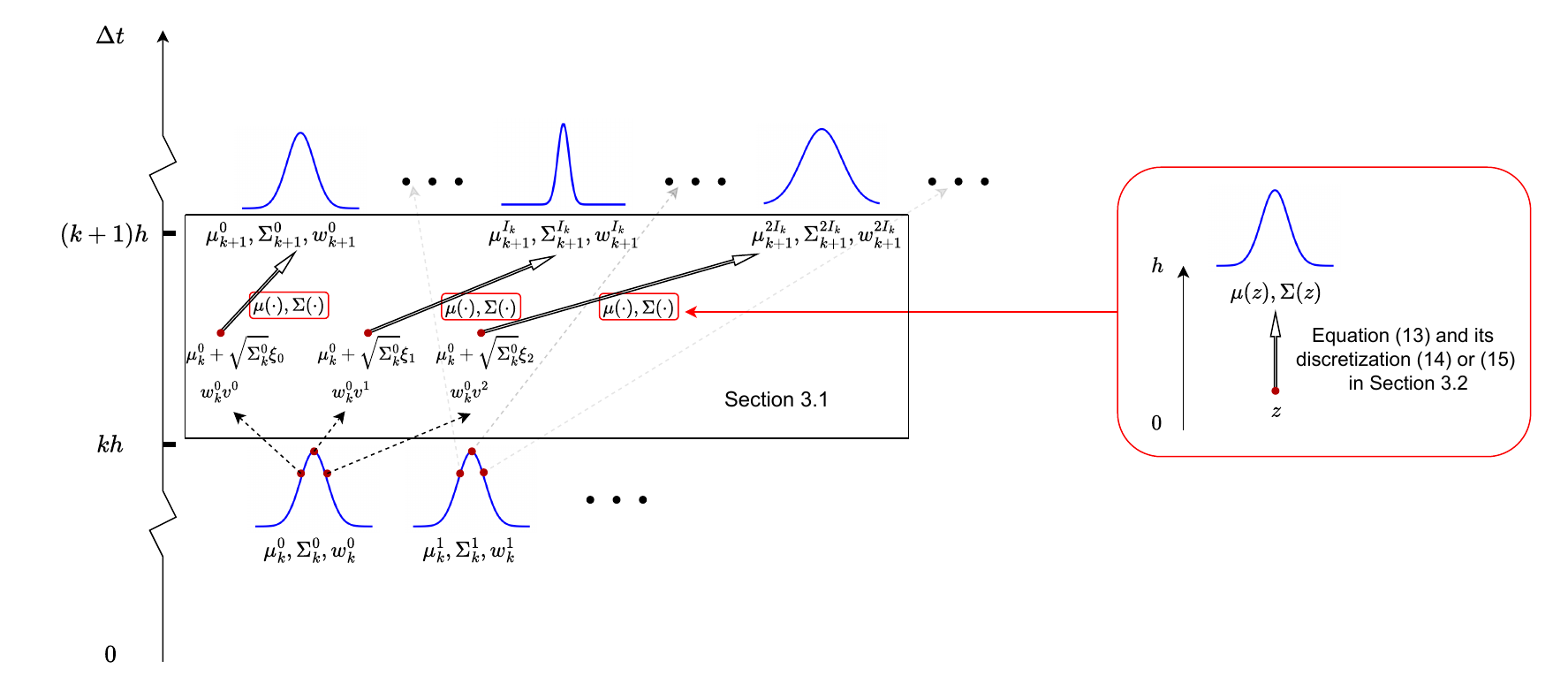}}
\vskip -0.5cm
\caption{Illustration of the proposed DynGMA with $3$ evaluation points. (Black box section) We discretize the time interval $[0, \Delta t]$ into $K$ sub-intervals with a step size of $h$, and then use Gaussian mixture model to approximate the $k$-step transition density (see \cref{DynGMA} for details). (Red box section) At each step of size $h$, we solve (\ref{eq: gauss discretization2}) to determine each Gaussian density within the Gaussian mixture model (see \cref{sec:gaussian} for details).}
    \label{fig:ill}
\end{figure}

Existing methods are usually based on one-step stochastic numerical schemes and thus necessitate data with sufficiently high time resolution. 
To make multiple time steps valid, thereby obtaining a more accurate approximation for large time intervals, we use the Gaussian mixture model to approximate the transition density $p(\Delta t, x(\Delta t)|0,  x(0); \theta)$ of SDE (\ref{eq:sdenet}) in this section. We refer to this approximation as dynamical Gaussian mixture approximation (DynGMA). An illustration of the proposed method is presented in \cref{fig:ill}.

First, we discretize the time interval $[0,\Delta t]$ into $K$ sub-intervals with a step size of $h$ such that $\Delta t = Kh$. At each step, we approximate the transition density by a Gaussian to compute, one after the other, approximations to the solutions $x(h), x(2h)$, $\cdots$, $x(Kh)$. The approximation of $x(kh)$ is denoted as $x_k$, and is obtained in the following recursive manner:
\begin{equation}\label{eq:gauss discretization}
x_{k+1} = \mu(x_k) + \sqrt{\Sigma(x_k)}\omega_{k+1},\quad k=0,\cdots, K-1,
\end{equation}
where $\omega_{k+1}\sim \N(\0_{D}, \1_{D\times D})$ is a $D$-dimensional standard normal distribution, $\mu(x_k) \in \R^D$, $\Sigma(x_k)\in \R^{D\times D}$ are the approximate mean and covariance of $x_{k+1}$, and $\sqrt{\Sigma(x_k)}$ is the Cholesky factorization of $\Sigma(x_k)$. This update formula (\ref{eq:gauss discretization}) yields the Gaussian approximation of the transition density $p((k+1)h, x_{k+1}|kh,  x_{k})$:
\begin{equation}\label{eq:one-step}
p_h((k+1)h, x_{k+1}|kh,  x_{k}) := \N(x_{k+1}|\mu(x_k), \Sigma(x_k)) \approx p((k+1)h, x_{k+1}|kh,  x_{k}).
\end{equation}
The computation of the Gaussian approximation, i.e., the explicit expressions of $\mu(\cdot)$ and $ \Sigma(\cdot)$, will be discussed in detail in next subsection.

Subsequently, using Chapman-Kolmogorov Equation \cite{pavliotis2016stochastic} and the Markov property of $\{x(kh)\}_{k=0}^K$, we are able to write the $k$-step transition density:
\begin{equation*}
\begin{aligned}
p((k+1)h, x_{k+1}|0, x_{0}) =& \int p((k+1)h, x_{k+1}|kh, x_{k}) p(kh, x_{k}|0, x_{0}) \d x_{k}.
\end{aligned}
\end{equation*}
This expression can be solved in a recursive manner by using approximation (\ref{eq:one-step}) but is still intractable due to the integration. Therefore, we adopt Gaussian mixture model as an approximation of each $k$-step transition density, i.e., 
\begin{equation*}
p_h(kh, x_{k}|0, x_{0}) := \sum_{i=0}^{I_k-1} w^i_{k} \N(x_{k}| \mu^i_k, \Sigma^i_{k})  \approx p(kh, x_{k}|0, x_{0}),
\end{equation*}
where $w^i_{k}$, $\mu^i_k$ and $\Sigma^i_{k}$ are parameters of which iterative formulas will be given later.
Then, we have
\begin{equation*}
\begin{aligned}
p((k+1)h, x_{k+1}|0, x_{0})\approx \sum_{i=0}^{I_k-1} w^i_{k}\int \N(x_{k+1}|\mu(x_k), \Sigma(x_k))  \N(x_{k}| \mu^i_k, \Sigma^i_{k}) \d x_{k},
\end{aligned}
\end{equation*}
which are some expectations with respect to the Gaussian density. Next, we employ cubature approximations to approximate the above integrals as weighted sums:
\begin{equation*}
\int \N(x_{k+1}|\mu(x_k), \Sigma(x_k))  \N(x_{k}| \mu^i_k, \Sigma^i_{k}) \d x_{k}
\approx \sum_{j=0}^J v^j \N(x_{k+1}|\mu(\mu^i_k + \sqrt{\Sigma^i_{k}}\vxi_j), \Sigma(\mu^i_k + \sqrt{\Sigma^i_{k}}\vxi_j)).
\end{equation*}
Here the coefficients $\vxi_j$ and $v^j$ determine the evaluation points and corresponding weights, respectively. There are general settings independent of specific problems, which will be provided in \cref{sec:Numerical experiments}. Subsequently, we obtain the Gaussian mixture approximation of $p((k+1)h, x_{k+1}|0, x_{0})$, yielding $p_h((k+1)h, x_{k+1}|0, x_{0})$:
\begin{equation*}
\begin{aligned}
p((k+1)h, x_{k+1}|0, x_{0})
\approx& \sum_{i=0}^{I_k-1} \sum_{j=0}^J w^i_{k} v^j \N(x_{k+1}|\mu(\mu^i_k + \sqrt{\Sigma^i_{k}}\vxi_j), \Sigma(\mu^i_k + \sqrt{\Sigma^i_{k}}\vxi_j))=: p_h((k+1)h, x_{k+1}|0, x_{0}).
\end{aligned}
\end{equation*}
The above definition of $p_h((k+1)h, x_{k+1}|0, x_{0})$ also yields recurrence relations for the parameters $w_k^i$, $\mu^i_k$ and $\Sigma^i_k$, i.e., 
\begin{equation*}
\begin{aligned}
\mu^i_{k+1} = & \mu(\mu^{(i \bmod I_k)}_k + \sqrt{\Sigma^{(i \bmod I_k)}_{k}}\vxi_{\lfloor i/I_k \rfloor}),\
&&\Sigma^i_{k+1} =  \Sigma(\mu^{(i \bmod I_k)}_k + \sqrt{\Sigma^{(i \bmod I_k)}_{k}}\vxi_{\lfloor i/I_k \rfloor}),\\
w^i_{k+1} = & w^{(i \bmod I_k)}_k v^{\lfloor i/I_k \rfloor},\  &&I_{k+1} =  (J+1) \cdot I_k,
\end{aligned}
\end{equation*}
which thus allows for the successive computation of the Gaussian mixture approximation.

\begin{remark}
It is worth noting that integrating multiple time discretization steps in density approximation is not trivial. For example, consider deriving density approximation via the Euler-Maruyama scheme,
\begin{equation*}
x_1 = x_0 + h f_{\theta}(x_0) + \sqrt{h} \sigma_{\theta}(x_0) \omega_1, \ x_2 = x_1 + h f_{\theta}(x_1) + \sqrt{h} \sigma_{\theta}(x_1) \omega_2,\ \cdots,\quad \omega_{k}\sim \N(\0_{D}, \1_{D\times D}),\ k=1,2,\cdots.
\end{equation*}
As discussed, $x_1$ is a Gaussian random variable, and its density can be expressed as (\ref{eq:GaussApp}). However, obtaining closed-form expressions for the densities of $x_k, k\geq 2$, is usually deemed intractable for general nonlinear $\sigma_{\theta}$.
\end{remark} 

\subsection{Gaussian density approximation via the asymptotic expansion of SDEs}\label{sec:gaussian}
\newcommand{\z}{z}
In this section, we derive the Gaussian approximation (9) employed in each sub-step. Due to the Markov property, it suffices to focus on the case where $k=0$.
As mentioned in the introduction, there are various Gaussian density approximations to the transition density. Since the error estimation of the existing Gaussian approximations is either lacking or based on one-step numerical schemes, in this section, we propose a novel Gaussian density approximation inspired directly by the random perturbation theory of dynamical systems \cite{blagoveshchenskii1962diffusion, blagoveshchenskii1961certain}. The proposed Gaussian density approximation is derived via a continuous ordinary differential equation. We have the flexibility to choose discretization methods (including adjusting the number of steps) in order to attain higher accuracy. Based on the random perturbation theory, we will also be able to quantify the error of the proposed Gaussian approximations and demonstrate the advantages of our method compared to the Euler-Maruyama approach (\ref{eq:GaussApp}). The proofs are postponed to \ref{app:proofs}. 

Next, we present the theorem of Blagoveshchenskii \cite{blagoveshchenskii1962diffusion}, which we further extend to learning SDEs using neural networks.
\begin{theorem}\label{thm:gauss0}
Consider an SDE with a small parameter:
\begin{equation*}
\d X(t) = f(X(t)) \d t + \varepsilon \sigma(X(t)) \d \omega(t),\quad  X (0) = \z.
\end{equation*}
Suppose elements of the drift $f$ and diffusion $\sigma$ have bounded partial derivatives up to order $2$, then there exists a Gaussian process $X_{GP}(t) = X_0(t) + \varepsilon X_1(t)$ such that
\begin{equation*}
\sup_{0\leq t \leq 1} \E{\norm{X(t) - X_{GP}(t)}_2^2} \leq C \varepsilon^4.
\end{equation*}
Here, $X_0(t)$ and $X_1(t)$ satisfy the following differential equations:
\begin{subequations}
\begin{align*}
\d X_0(t) &= f (X_0(t)) \d t, &&  X_0(0) = \z,\\
\d X_1(t) &= J_{f } (X_0(t)) X_1(t) \d t +  \sigma (X_0(t)) \d \omega(t),\quad  &&  X_1(0) = \0_{D},
\end{align*}
\end{subequations}
where $J_{f } = \frac{\partial f }{\partial x}$ denotes the Jacobian of $f$, and $\E{\cdot}$ denotes the expectation over the Brownian motion.
\end{theorem}

\cref{thm:gauss0} indicates the feasibility of Gaussian density to approximate the transition density, and leads to the following corollary for learning SDEs using neural networks immediately.
\begin{corollary}\label{cor:gauss}
Consider a parameterized SDE 
\begin{equation}\label{eq:sdenetk}
\d x(t) = f_{\theta}(x(t)) \d t + \sigma_{\theta}(x(t)) d \omega(t), \quad x(0)=\z.
\end{equation}
Suppose elements of $f_{\theta}$ and  $\sigma_{\theta}/\norm{\sigma_{\theta}}_F$ in (\ref{eq:sdenetk}) have bounded partial derivatives up to order $2$, then there exists a Gaussian random variable $X_G \sim \N(\mu, \Sigma)$ such that
\begin{equation*}
\E{\norm{x(h) - X_G}_2^2} \leq C h^2\norm{\sigma_{\theta}}_F^4.
\end{equation*}
Here, $\norm{\sigma_{\theta}}_F=\sup_{x\in\R^D}\norm{\sigma_{\theta}(x)}_F$, $\mu = X_0(h)$, $\Sigma = \E{X_1(h)X_1^{\top}(h)}$, and $X_0(t)$,  $X_1(t)$ obey differential equations:
\begin{subequations}
\begin{align}
\d X_0(t) &= f_{\theta} (X_0(t)) \d t, &&  X_0(0) = \z,\label{eq:mean}\\
\d X_1(t) &= J_{f_{\theta}} (X_0(t)) X_1(t) \d t + \sigma_{\theta}(X_0(t)) \d \omega(t),\quad  &&  X_1(0) = \0_{D},\label{eq:X2}
\end{align}
\end{subequations}
where $J_{f_{\theta}} = \frac{\partial f_{\theta}}{\partial x}$ represents the Jacobian of $f_{\theta}$, and $\E{\cdot}$ denotes the expectation over the Brownian motion.
\end{corollary}
Equation (\ref{eq:mean}) is a deterministic system determining the mean of $X_G$, and can be solved numerically. The solution of the SDE (\ref{eq:X2}) is a Gaussian process with a zero mean determining the covariance of $X_G$. If $X_0(t)$ is known, by FPK equation, the covariance $\NSigma(t)$ of the solution to the SDE (\ref{eq:X2}) are given by
\begin{equation*}
\begin{aligned}
\frac{\d \NSigma(t)}{\d t} =& \int \left(J_{f_{\theta}} (X_0(t)) x x^{\top} + x  x^{\top}  J_{f_{\theta}} (X_0(t)) + \sigma_{\theta}(X_0(t)) \sigma_{\theta}^{\top}(X_0(t)) \right)p(x,t) \d x,
\end{aligned}
\end{equation*}
where $p(x,t)$ is the density of $ X_1(t)$. The mean $\mu$ and covariance $\Sigma$ of Gaussian random variable $X_G$ issued from $\z$ can be viewed as a function with respect to $\z$. 
Observing that $\int x  x^{\top} p(x,t) \d x = \Sigma(t)$ (since the mean of $X_1$ is zero), and combining equation (\ref{eq:mean}), they are written as
\begin{equation}\label{eq: gauss discretization2}
\begin{aligned}
\frac{\d \Nmu (t)}{\d t} =& f_{\theta}(\Nmu(t)), \quad &&\Nmu(0)=\z,\quad &&&\mu(\z)=\Nmu(h),\\
\frac{\d \NSigma(t)}{\d t} =& J_{f_{\theta}}(\Nmu(t)) \NSigma(t) + \NSigma(t)  J_{f_{\theta}}^{\top}(\Nmu(t)) +  \sigma_{\theta}(\Nmu(t))\sigma_{\theta}^{\top}(\Nmu(t)), \quad &&\NSigma(0)=\0_{D\times D}, &&&\Sigma(\z)=\NSigma(h).
\end{aligned}
\end{equation}
Subsequently, we discretize the time interval $[0, h] $ into $L$ sub-intervals and design the following iteration formula to obtain the approximations of $\mu(\z)$ and $\Sigma(\z)$:
\begin{equation}\label{eq:ite of con}
\begin{aligned}
&\Nmu_0 = \Nmu(0) =\z,\quad \NSigma_0 = \NSigma(0)=\0_{D\times D};\\
&\Nmu_{l+1/2} = \Nmu_{l} + \frac{h}{2L} f_{\theta}(\Nmu_{l}), \quad
\Nmu_{l+1} = \Nmu_{l} + \frac{h}{L} f_{\theta}(\Nmu_{l+1/2});\\
&\NSigma_{l+1} = \left(\1_{D\times D} + \frac{h}{L}  J_{f_{\theta}}(\Nmu_{l+1/2})\right) \NSigma_{l} \left(\1_{D\times D} + \frac{h}{L}  J_{f_{\theta}}(\Nmu_{l+1/2})\right)^{\top}\\
&\quad\quad\quad +\frac{h}{L} \left(\1_{D\times D} + \frac{h}{2L}  J_{f_{\theta}}(\Nmu_{l+1/2})\right) \sigma_{\theta}(\Nmu_{l+1/2})\sigma_{\theta}^{\top}(\Nmu_{l+1/2}) \left(\1_{D\times D} + \frac{h}{2L}  J_{f_{\theta}}(\Nmu_{l+1/2})\right)^{\top};\\
&\mu(\z) := \Nmu_L,\quad \Sigma(\z) := \NSigma_L.
\end{aligned}
\end{equation}
Then, the Gaussian approximation in \cref{DynGMA} is given as
\begin{equation*} 
p_h((k+1)h, x_{k+1}|kh,  x_{k}) := \N(x_{k+1}|\mu(x_k), \Sigma(x_k)),
\end{equation*}
where $\mu(\cdot)$, $\Sigma(\cdot)$ are defined by (\ref{eq:ite of con}). The complete DynGMA using discretization (\ref{eq:ite of con}) is summarized in \cref{alg:dyngma1}.

\begin{algorithm}
\renewcommand{\algorithmicrequire}{\textbf{Input:}}
\renewcommand{\algorithmicensure}{\textbf{Output:}}
\caption{Dynamical Gaussian mixture approximation for SDE}\label{alg:dyngma1}
\begin{algorithmic}
\State Choose parameters of cubature, $v^j$, $\xi_j$, $j=0,\cdots, J$.
\State Input time step $\Delta t \geq 0$, neural networks $f_{\theta}$ and $\sigma_{\theta}$.
\State Input initial mean and covariance $\mu_0^0$ and $\Sigma_0^0$ of $x_0$. If $x_0$ is a single point, then $\mu_0^0=x_0$ and $\Sigma_0^0=0$.
\State $I_0=1$, $w_{0}^0=1$.
\State Divide the interval $[0,\ \Delta t]$ to $K$ subintervals of length $h$.

\For{${0}\leq k \leq K-1$}
\For{$0\leq i \leq I_{k}-1$}
\State Compute the Cholesky factorization of $\Sigma^{i}_{k}$.
\EndFor
\State $I_{k+1} = (J+1) I_k$.
\For{$0\leq i \leq I_{k+1}-1$}
\State $w^i_{k+1} = w^{(i \bmod I_k)}_k v^{\lfloor i/I_k \rfloor}$.

\State $x_k= \mu^{(i \bmod I_k)}_k + \sqrt{\Sigma^{(i \bmod I_k)}_{k}}\vxi_{\lfloor i/I_k \rfloor}$.

\State $\mu^i_{k+1} =  \mu(x_k)$, $\Sigma^i_{k+1} =  \Sigma(x_k)$ according to (\ref{eq:ite of con}).
\EndFor
\EndFor
\State \Return $p_{h}(\Delta t_, x_K|0,  x_0; \theta) = \sum_{i=0}^{I_K-1} w_{K}^i \N(x_K|\mu_K^i, \Sigma_K^i)$.
\end{algorithmic}
\end{algorithm}

\begin{algorithm}
\renewcommand{\algorithmicrequire}{\textbf{Input:}}
\renewcommand{\algorithmicensure}{\textbf{Output:}}
\caption{Dynamical Gaussian mixture approximation for SDE without Cholesky factorization}\label{alg:dyngma2}
\begin{algorithmic}
\State Choose parameters of cubature, $v^j$, $\xi_j$, $j=0,\cdots, J$.
\State Input time step $\Delta t \geq 0$, neural networks $f_{\theta}$ and $\sigma_{\theta}$.
\State {Input initial mean and covariance $\mu_0^0$ and $\Sigma_0^0$ of $x_0$. If $x_0$ is a single point, then $\mu_0^0=x_0$ and $\Sigma_0^0=0$.}
\State $I_0=1$, $w_{0}^0=1$.
\State Divide the interval $[0,\ \Delta t]$ to $K$ subintervals of length $h$.

\For{${0}\leq k \leq K-1$}
\State $I_{k+1} = (J+1) I_k$.
\For{$0\leq i \leq I_{k+1}-1$}
\State $w^i_{k+1} = w^{(i \bmod I_k)}_k v^{\lfloor i/I_k \rfloor}$.

\State $x_k= \mu^{(i \bmod I_k)}_k + \sqrt{\Sigma^{(i \bmod I_k)}_{k}}\vxi_{\lfloor i/I_k \rfloor}$.

\State $\mu^i_{k+1}=\mu(x_k)$, $\sqrt{\Sigma^i_{k+1}}=\sqrt{\Sigma(x_k)}$  according to (\ref{eq:ite of den cov2}).
\EndFor
\EndFor
\State \Return $p_{h}(\Delta t_, x_K|0,  x_0; \theta) = \sum_{i=0}^{I_K-1} w_{K}^i \N(x_K|\mu_K^i, \sqrt{\Sigma_K^i}\sqrt{\Sigma_K^i}^{\top})$.
\end{algorithmic}
\end{algorithm}

Combining classical numerical analysis theory, the Gaussian approximation using discretization (\ref{eq:ite of con}) has the following properties:
\begin{theorem}\label{the:dis con}
Under conditions of \cref{cor:gauss}, and further assume that the diffusion matrix $\sigma_{\theta}\sigma_{\theta}^{\top}$ is strictly positive definite, i.e., $\sum_{i,j} [\sigma_{\theta}(x)\sigma_{\theta}^{\top}(x)]_{ij} \lambda_i \lambda_j \geq C'\sum_i \lambda_i^2$ for any real $\lambda_1,\cdots, \lambda_D$ and $x\in \R^D$, where $C'$ is a positive constant. There
exists a Gaussian random variable $X_h \sim \N(\mu(\z), \Sigma(\z))$ where $\mu(\z)$ and $\Sigma(\z)$ are defined in (\ref{eq:ite of con})
and $\Sigma(\z)$ is strictly positive definite, such that
\begin{equation*}
\E{\norm{x(h) - X_h}_2^2} \leq C h^2(\norm{\sigma_{\theta}}_F^4 + \norm{\sigma_{\theta}}_F^2 h^3/L^4 + h^4/L^4).
\end{equation*}
\end{theorem}
It is noted that the matrix $\Sigma(\z)$ requires Cholesky factorization, and its determinant will serve as the denominator in the MLE. Therefore, its strict positivity is essential for ensuring stable training progress. Moreover, the SDEs of interest arise from perturbed dynamical systems, where the magnitude of the diffusion term, i.e., $\norm{\sigma_{\theta}}_F$, is typically small. Consequently, the leading term in the error bound is the last term, $Ch^6/L^4$.

Subsequently, we propose an alternative discretization of (\ref{eq: gauss discretization2}) to circumvent Cholesky factorization by combining the parameterization discussed in \cref{ssec:Parameterization}. We use the following one-step scheme to discrete (\ref{eq: gauss discretization2}):
\begin{equation}\label{eq:ite of den cov2}
\begin{aligned}
&\Nmu_{1/2} = \z + \frac{h}{2} f_{\theta}(\z), \quad
\mu(\z) = \z + h f_{\theta}(\Nmu_{1/2}),\\
&\sqrt{\Sigma(\z)} = \sqrt{h} \sigma_{\theta}(\Nmu_{1/2}).
\end{aligned}
\end{equation}
Here $\sqrt{\Sigma(\z)}$ is the Cholesky factorization of $h \sigma_{\theta}(\Nmu_{1/2})\sigma_{\theta}^{\top}(\Nmu_{1/2})$ if $\sigma_{\theta}(\Nmu_{1/2})$ is a lower triangular matrix, thereby circumventing the computation of Cholesky factorization in DynGMA. The complete algorithm using (\ref{eq:ite of den cov2}) to discretize (\ref{eq: gauss discretization2}) is summarized in \cref{alg:dyngma2}. Similarly, we have the following estimate:
\begin{theorem}\label{the:dis con2}
Under conditions of \cref{the:dis con}, there exists a Gaussian random variable
$X_h \sim \N(\mu(\z), \Sigma(\z))$ where $\mu(\z)$ and $\Sigma(\z)$ are defined in (\ref{eq:ite of den cov2}), such that
\begin{equation*}
\E{\norm{x(h) - X_h}_2^2} \leq C h^2(\norm{\sigma_{\theta}}_F^4 + \norm{\sigma_{\theta}}_F^2 h  + h^4 ).
\end{equation*}
\end{theorem}

The proposed DynGMA is closely related to the Euler-Maruyama approach \cite{dietrich2023learning}. In the case when there is solely one Gaussian mixture step (i.e., $K=1$), and the selected discretization scheme for (\ref{eq: gauss discretization2}) is the one-step Euler method, DynGMA transforms into a Gaussian approximation (\ref{eq:GaussApp}) derived through the Euler-Maruyama approach. According to the random perturbation theory of dynamical systems, the estimates for (\ref{eq:GaussApp}) can be written as follows:
\begin{theorem}\label{the:dis con3}
Under conditions of \cref{the:dis con}, there exists a Gaussian random variable
$X_h \sim \N(\mu(\z), \Sigma(\z))$ where $\mu(\z) = \z + h f_{\theta}(\z)$ and $\Sigma(\z) = h \sigma_{\theta}(\z) \sigma^{\top}_{\theta}(\z)$, such that
\begin{equation*}
\E{\norm{x(h) - X_h}_2^2} \leq C h^2(\norm{\sigma_{\theta}}_F^4 + \norm{\sigma_{\theta}}_F^2 h  + h^2).
\end{equation*}
\end{theorem} 
We should point out that the Euler-Maruyama approach is highly efficient but has low accuracy. The proposed DynGMA offers a flexible approach. We are able to select discretization schemes of (\ref{eq: gauss discretization2}), such as (\ref{eq:ite of con}) and (\ref{eq:ite of den cov2}), based on practical requirements to achieve a balance between computational efficiency and accuracy. \cref{alg:dyngma1} requires more steps to achieve higher accuracy, whereas \cref{alg:dyngma2} trades a certain level of accuracy for higher efficiency. In particular, the proposed DynGMA, implemented with \cref{alg:dyngma2} and $K=1$, involves only one additional function evaluation yet enhances the accuracy by one order for small random perturbations, compared to the Euler-Maruyama approach. 
In practical applications, we can primarily use the single-step Gaussian approximation, i.e., $K=1$, as it is not only theoretically valid but also offers higher accuracy. The mixed model, i.e., $K>1$, is employed as a supplement to improve accuracy with larger step sizes.

\subsection{Parameterization and loss function}\label{ssec:Parameterization}
We use neural networks to parameterize $f_{\theta}$ and $\sigma_{\theta}$ in (\ref{eq:sdenet}). A feed-forward fully-connected neural network (FNN) is composed of a series of linear layers and activation layers, i.e.,
\begin{equation*}
\begin{aligned}
\mathcal{NN}(x) = W_{S+1} h_S\circ h_{S-1}\circ \cdots \circ h_1(x) +b_{S+1},
\end{aligned}
\end{equation*}
where
\begin{equation*}
h_s(z_s) =\eta (W_sz_s+b_s),\ W_s\in \R^{d_s \times d_{s-1}},\ b_s\in \R^{d_s} \text{ for }s=1,\cdots,S+1.
\end{equation*}
The number $S$ is the total number of layers, and $\{W_s, b_s\}_{s=1}^{S+1}$ denotes all the trainable parameters. $\eta:\R \rightarrow \R$ is a predetermined activation function that is applied element-wise to a vector. 
In this paper, we employ the hyperbolic tangent function, denoted as $\texttt{tanh}(z) = (e^z - e^{-z})/(e^z + e^{-z})$. 
Specifically, we parameterize $f_{\theta}$ by an FNN. For state-dependent diffusion, we parameterize $\sigma_{\theta}$ by the combination of a strictly lower triangular matrix and a positive diagonal matrix:
\begin{equation}\label{eq:sigma}
\sigma_{\theta}(x) = \sigma_{\theta,1}(x) + \text{diag}\left(\frac{\sqrt{\sigma^2_{\theta,2}(x)+1}+\sigma_{\theta,2}(x)}{2}\right),
\end{equation}
where $\sigma_{\theta,1}(x) \in \R^{D\times D}$, $[\sigma_{\theta,1}(x)]_{ij}=0$ for $i\leq j$ is obtained by rearranging the elements of an FNN with an output dimension of $D^2$, and $\sigma_{\theta,2}(x) \in \R^D$ is directly produced by an FNN. For constant diffusion, we parameterize $\sigma_{\theta}$ with a $D$-by-$D$ matrix if using \cref{alg:dyngma1}, and with a strictly lower triangular constant matrix by employing constant matrices $\sigma_{\theta,1}$ and $\sigma_{\theta,2}$ in (\ref{eq:sigma}) if using \cref{alg:dyngma2}. We denote all the trainable parameters of these three FNNs as $\theta$. It is noted that the proposed parameterization of $\sigma_{\theta}$ is still able to model any positive definite diffusion matrices due to Cholesky factorization, which states that any real positive definite matrix can be decomposed into the product of a lower triangular matrix and its transpose. 

Given training trajectories of the form (\ref{eq:data}), we can then solve the MLE (\ref{eq:mle}) by optimizing the likelihood loss:
\begin{equation}\label{eq:loss}
\begin{aligned}
\text{Loss}(\theta) = -\frac{1}{NM}\sum_{n=1}^N\sum_{m=1}^M \log p_h(\Delta t_m, y^n_m|0,  y^n_{m-1}; \theta),
\end{aligned}
\end{equation}
where $p_h$ is the DynGMA to the transition density of the parameterized SDE discussed above.
Since 
\begin{equation*}
\N(x|\mu, \Sigma) = \frac{1}{(2\pi)^{D/2} \det(\Sigma)^{1/2}}e^{-\frac{1}{2} (x-\mu)^{\top}\Sigma^{-1}(x-\mu)},
\end{equation*}
the proposed approximation $p_{h}(\Delta t, x_K|0,  x_0; \theta)$ has a smooth representation, and thus has forward and backward implementations in established automatic differentiation packages. We can choose to optimize the loss using stochastic gradient-based optimization methods such as Adam optimization \cite{kingma2014adam}. Alternatively, we propose multi-step training with DynGMA by optimizing the likelihood loss:
\begin{equation}\label{eq:mloss}
\text{Loss}(\theta) = -\frac{1}{N\lfloor M/\gamma\rfloor}\sum_{\substack{\gamma \in \Gamma\\ \Gamma \subset \{1, \cdots, M\}}}\sum_{n=1}^N\sum_{m=1}^{\lfloor M/\gamma\rfloor} \log p_{h}(\Delta t_{m,\gamma}, y^{n}_{m\gamma}|0,  y^{n}_{(m-1)\gamma}; \theta),
\end{equation}
which is obtained by (\ref{multi-steploss}). 
In the next section, we will numerically demonstrate that multi-step training enables DynGMA to handle data with measurement noise effectively.

\section{Numerical experiments}\label{sec:Numerical experiments}
In this section, we first show that the proposed DynGMA is able to accurately approximate the transition density on Bene\v{s} SDE. Then we show its benefits in learning SDEs on several widely investigated benchmark problems, including a two-dimensional ($2$d) system, the Lorenz system ($3$d), a practical biological model ($10$d), and the Susceptible-Infectious-Recovered model ($2$d). We will also demonstrate the advantages of the proposed method in predicting dynamics and computing invariant distributions.

The baseline method used for comparison in this section is the Gaussian approximation (\ref{eq:GaussApp}) obtained via the Euler-Maruyama \cite{dietrich2023learning}, of which effectiveness has been illustrated in various applications such as learning effective SDEs from microscopic simulations \cite{dietrich2023learning}, computing invariant distributions from data \cite{lin2023computing}, and learning macroscopic thermodynamic description of stretching dynamics of polymer chains \cite{chen2023constructing}. We employ a similar parameterization for the baseline method, and its loss function is obtained by substituting (\ref{eq:GaussApp}) into (\ref{eq:loss}), as follows:
\begin{equation*}
\begin{aligned}
&\text{Loss}(\theta) = \frac{1}{2NM}\sum_{n=1}^N\sum_{m=1}^M \Big[(y^n_m - \mu_{em}(y^n_{m-1}))^{\top} \Sigma_{em}^{-1}(y^n_{m-1})(y^n_m - \mu_{em}(y^n_{m-1})) +  \log \det \Sigma_{em}(y^n_{m-1})  + D \log 2\pi\Big],
\end{aligned}
\end{equation*}
where
\begin{equation*}
\Sigma_{em}(y^n_{m-1}) = \Delta t_m \sigma_{\theta}(y^n_{m-1}) \sigma^{\top}_{\theta}(y^n_{m-1}), \quad \mu_{em}(y^n_{m-1}) = y^n_{m-1} + \Delta t_m f_{\theta}(y^n_{m-1}).
\end{equation*}
We refer to this approach as the Euler-Maruyama approach. Another method for comparison is the Gaussian approximation obtained via the FPK equation and the cubature approximation \cite{kushner1967approximations, sarkka2013gaussian, solin2021scalable}. The corresponding loss function is of the form
\begin{equation*}
\begin{aligned}
&\text{Loss}(\theta) = \frac{1}{2NM}\sum_{n=1}^N\sum_{m=1}^M \Big[(y^n_m - \mu_{cub}(y^n_{m-1}))^{\top} \Sigma_{cub}^{-1}(y^n_{m-1})(y^n_m - \mu_{cub}(y^n_{m-1})) +  \log \det \Sigma_{cub}(y^n_{m-1})  + D \log 2\pi\Big],
\end{aligned}
\end{equation*}
where $\mu_{cub}(y^n_{m-1}) = \nmu(\Delta t_m)$, $\Sigma_{cub}(y^n_{m-1}) = \nSigma(\Delta t_m)$, and $\nmu$, $\nSigma$ are obtained by numerically solving the ODE:
\begin{equation*}
\begin{aligned}
\frac{\d \nmu}{\d t} = &  \sum v^j f_{\theta}(\nmu +  \sqrt{\nSigma} \vxi_j), \quad &&  \nmu(0)=y^n_{m-1}\\
\frac{\d \nSigma}{\d t} = & \sum v^j f_{\theta}(\nmu +  \sqrt{\nSigma} \vxi_j) \vxi_j^{\top}\sqrt{\nSigma}^{\top}  + \sum v^j \sqrt{\nSigma} \vxi_j f_{\theta}(\nmu +  \sqrt{\nSigma} \vxi_j)^{\top} \\
&+ \sum v^j \sigma_{\theta}(\nmu +  \sqrt{\nSigma} \vxi_j, t)\sigma_{\theta}^{\top}(\nmu +  \sqrt{\nSigma} \vxi_j, t), &&\nSigma(0)=0.
\end{aligned}
\end{equation*}
We refer to this approach as the Gaussian Cubature approach. The computational complexity of both the Gaussian Cubature approach and DynGMA heavily relies on the selection of coefficients $\xi_j$ and $v^j$. A commonly chosen method is the symmetric cubature rule obtained through the unscented transform \cite{arasaratnam2009cubature,mcnamee1967construction, sarkka2023bayesian}, which precisely encodes the given mean and covariance information in the selected set of points:
\begin{equation*}
\begin{aligned}
\vxi_0=0,\quad 
\vxi_j = \left\{\begin{aligned}
&\sqrt{D+1}e_j, \quad && j=1, \cdots, D,\\
 - &\sqrt{D+1}e_j, \quad && j=D+1, \cdots, 2D,\\
\end{aligned}\right. 
\qquad v^0=\frac{1}{D+1},\quad  v^j = \frac{1}{2(D+1)},
\end{aligned}
\end{equation*}
where $e_j$ is the unit coordinate vector. This rule can be constructed up to any desired order \cite{mcnamee1967construction}. Given our focus on dynamical systems with random perturbations, where the variance magnitude is typically small, we adopt the aforementioned parameter settings for both the Gaussian Cubature approach and DynGMA in this paper.

If not otherwise specified, we use the Euler-Maruyama scheme with very fine step size\footnote{Specifically, $\Delta t/10$ in \cref{ssec:2d}, $\Delta t/20$ in \cref{ssec:ls}, $5\times 10^{-4}$ in \cref{ssec:emt}, and $10^{-4}$ in \cref{ssec:sir}.} to simulate the underlying stochastic dynamics for generating data and to simulate the learned systems for numerical verification. Moreover, we compute the average relative error to provide a quantitative evaluation of the performance
\begin{equation*}
e_f = \brac{\frac{\sum_{x\in \mathcal{T}} \norm{f_{\theta}(x) - f(x)}_2^2}{\sum_{x\in \mathcal{T}} \norm{f(x)}_2^2}}^{1/2},\quad e_{\sigma} = \brac{\frac{\sum_{x\in \mathcal{T}} \norm{\sigma_{\theta}(x) - \sigma(x)}_2^2}{\sum_{x\in \mathcal{T}} \norm{\sigma(x)}_2^2}}^{1/2},
\end{equation*}
where $\mathcal{T}$ is composed of several points randomly sampled from the given domain. And we compute $e_P$ and $e_V$ in a similar way, where $P$ represents the invariant distribution and $V$ is the generalized potential, $V = - \norm{\sigma}_F^2/2 \log P$.
All neural network models are trained on the PyTorch framework, and the code accompanying the numerical experiments is publicly available at \url{https://github.com/Aiqing-Zhu/DynGMA}.

\subsection{Improving accuracy of density approximation}
In this subsection, we show that the proposed DynGMA and the Gaussian density approximation via the asymptotic expansion can yield better density approximations compared with the Euler-Maruyama approach. Consider the Bene\v{s} SDE:
\begin{equation*}
\d x = \tanh x \d t + \d \omega,
\end{equation*}
where $\omega$ is a standard Brownian motion. For any initial value $x_0$, the transition  density is given as
\begin{equation*}
p(t, x | 0, x_0) = \frac{1}{\sqrt{2\pi t}} \frac{\cosh x }{\cosh x_0} \exp \brac{-\frac{1}{2}t} \exp\brac{-\frac{1}{2t}(x  - x_0)^2}.
\end{equation*}
\begin{figure}[htbp]
    \centering
    \centerline{\includegraphics[width=1\linewidth]{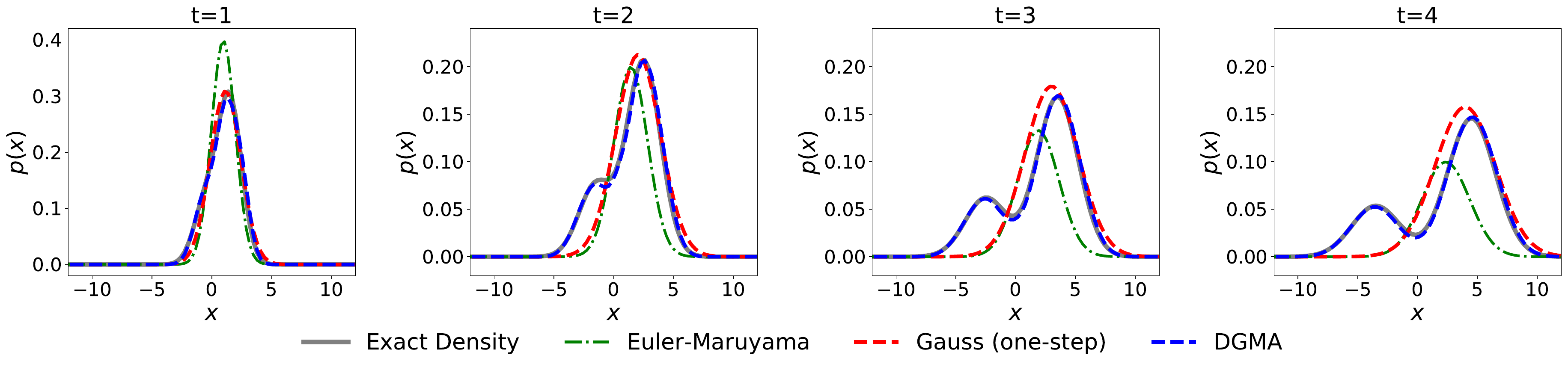}}
    \caption{Comparison of approximations and the exact density for Bene\v{s} SDE.}
    \label{fig:bene}
\end{figure}
Here we consider three kinds of approximations for the exact density: (i) the one-step Euler-Maruyama approach, i.e., Gaussian approximation with mean $x_0 + t \tanh x_0$ and covariance $\sqrt{t}$; (ii) the one-step Gaussian approximation , i.e., numerically solving Equation (\ref{eq: gauss discretization2}) using (\ref{eq:ite of con}) with $L=2$ to obtain the mean and covariance; and (iii) the proposed DynGMA using \cref{alg:dyngma1}  with parameter settings specified as $h=0.5$ corresponding to $K=t/0.5$ and $L=2$.

With an initial condition of $x_0 = 1/2$, the comparison of these approximations to the exact density is illustrated in \cref{fig:bene}. As depicted in the figure, the proposed DynGMA demonstrates a significantly more accurate approximation, especially noticeable when the time step is large. Furthermore, considering that the one-step Euler-Maruyama scheme is inherently a Gaussian approximation, the leftmost figure highlights that the proposed Gaussian approximation yields more accurate mean and covariance results. These findings demonstrate that numerically solving Equation (\ref{eq: gauss discretization2}) with utilizing (\ref{eq:ite of con}), instead of using the Euler-Maruyama, in each sub-step to obtain the Gaussian approximation within the proposed DynGMA framework, facilitates a more accurate approximation of the transition density. Furthermore, the proposed DynGMA is also capable of learning more accurate approximation of the drift and diffusion functions from data. Experimental results and details are presented in \ref{app:Learning Benes SDE}.

\subsection{Improving accuracy of learning SDEs}\label{ssec:2d}
In this subsection, we demonstrate that, compared to the Euler-Maruyama approach, learning SDEs using DynGMA (i) exhibits superior accuracy with a given step size; (ii) can handle larger step sizes effectively; and (iii) performs well even in the presence of higher levels of measurement noise.
As the first illustration of learning SDEs, we examine a two-dimensional system that has been studied for computing invariant distributions from data \cite{lin2023computing}. In that research, the Euler-Maruyama approach is employed to learn the governing functions, and the invariant distribution is computed simultaneously. Here, we use the proposed DynGMA to reconstruct the entire dynamics and then compute the invariant distribution using their decomposition method. The system is formulated as:
\begin{equation}\label{eq:2d}
\begin{aligned}
\d x  &= \brac{ x(1-x^2)/5 + y(1+\sin x)}\d t + \sqrt{\frac{1}{50}} \d \omega_1,\\
\d y  &= \brac{-y + 2x(1-x^2)(1+\sin x)}\d t + \sqrt{\frac{1}{5}} \d \omega_2,
\end{aligned}
\end{equation}
where $\omega_1$ and $\omega_2$ are independent Brownian motions.

We generated $2000 \left(\frac{\Delta t}{0.05}\right)$ trajectories for our training data. Each trajectory consists of an initial state uniformly sampled from $[-2,2]\times[-3,3]$, and includes $M=1/\Delta t$ states at equidistant time steps of $\Delta t$. In other words, there are a total of $4\times 10^4$ data points in the dataset. We employed an FNN with two hidden layers, each having 128 units, and using tanh activation function to parameterize the drift $f_{\theta}$. Additionally, We use a constant $2$-by-$2$ matrix to parameterize the diffusion $\sigma_{\theta}$. 
We use Algorithm 1 to approximate the transition density. Here the step size of each sub-Gaussian step is set to $h=0.1$ and the step size of discretization (\ref{eq:ite of con}) is fixed at $0.05$. The number of steps is determined as $K=\lceil \Delta t / h \rceil$ and $L=\min(h, \Delta t)/0.05$. Specifically, for $\Delta t$ values of $0.05$, $0.1$, and $0.2$, the values of $K$ are $1, 1, 2$, and the values of $L$ are $1, 2, 2$. 
During training, we employed full-batch Adam optimization for $5\times 10^3$ epochs, with the learning rate set to exponentially decay with linearly decreasing powers from $10^{-2}$ to $10^{-4}$.

\begin{figure}[!ht]
\centerline{\includegraphics[width=1\linewidth]{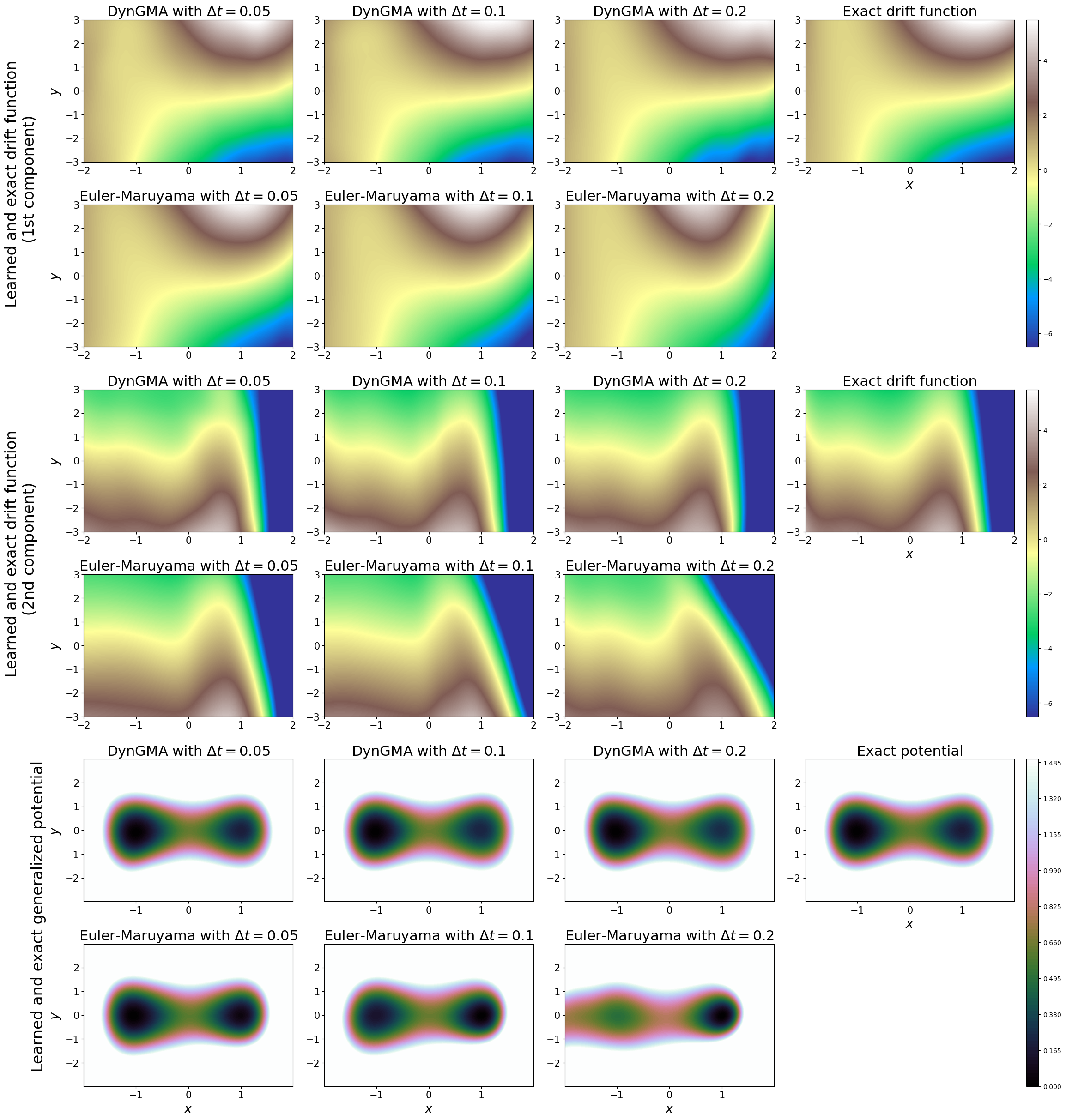}}
\caption{Learned drift functions and generalized potential of the learned dynamics  for the two-dimensional system (\ref{eq:2d}).}
    \label{fig:2d}
\end{figure}
\begin{figure}[!ht]
\centerline{\includegraphics[width=1\linewidth]{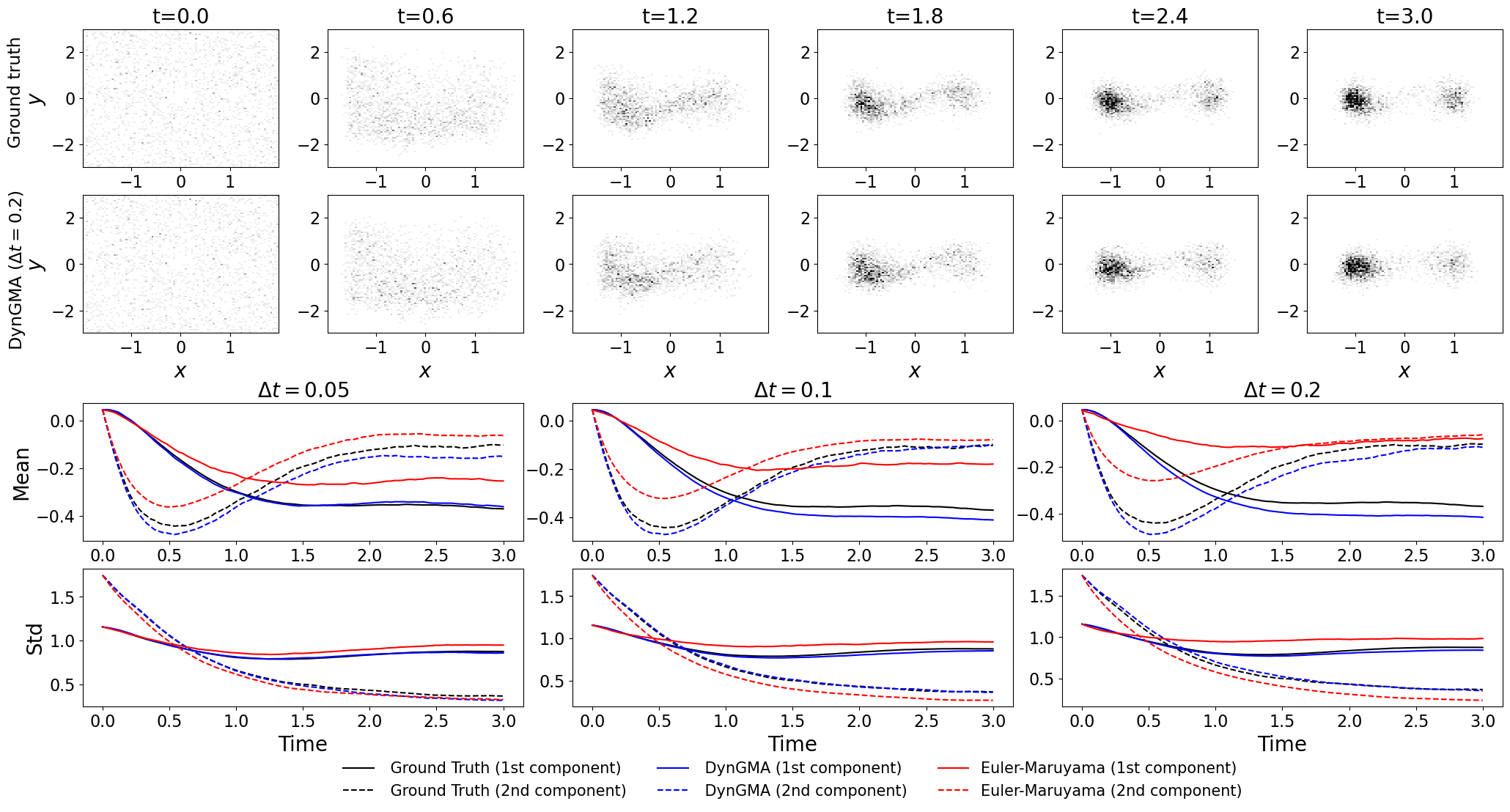}}
\caption{Time evolution of the distribution (Top), the mean and the standard deviation (Bottom) of the learned and exact dynamics for the two-dimensional system.}
    \label{fig:2d_traj}
\end{figure}
\begin{table}[ht]
  \centering
    \resizebox{\textwidth}{!}{
    \begin{tabular}{ccccccc}
    \toprule
    \multirow{2}{*}{Settings}&
    \multicolumn{2}{c}{$\Delta t=0.05$}&\multicolumn{2}{c}{$\Delta t=0.1$}&\multicolumn{2}{c}{$\Delta t=0.2$}\cr
    \cmidrule(lr){2-3} \cmidrule(lr){4-5} \cmidrule(lr){6-7}
    & DynGMA&Euler-Maruyama& DynGMA&Euler-Maruyama& DynGMA&Euler-Maruyama\cr
    \midrule
    $e_{\sigma}$ 
& 9.94e-3$\pm$4.13e-3 & 4.05e-2$\pm$4.49e-3 & 5.21e-3$\pm$2.73e-3 & 9.22e-2$\pm$1.85e-3 & 3.77e-2$\pm$5.27e-3 & 1.58e-1$\pm$4.37e-3 \cr
$e_f$
& 4.59e-2$\pm$4.71e-3 & 1.55e-1$\pm$4.66e-3 & 4.38e-2$\pm$2.39e-3 & 2.93e-1$\pm$3.84e-3 & 7.45e-2$\pm$1.20e-3 & 4.96e-1$\pm$6.11e-3 \cr
$e_V$
& 5.33e-2$\pm$4.64e-2 & 1.26e-1$\pm$1.50e-2 & 7.63e-2$\pm$3.55e-2 & 2.29e-1$\pm$6.77e-2 & 8.70e-2$\pm$7.82e-2 & 4.72e-1$\pm$3.73e-2 \cr
    \bottomrule
    \end{tabular} 
    }
    \caption{Quantitative results of the two-dimensional system (\ref{eq:2d}). The results are obtained by taking the mean of $5$ independent experiments, and the errors are recorded in the form of mean $\pm$ standard deviation.}
    \label{tab:error2d}
\end{table}

After completing the training process, we compute the invariant distributions and the generalized potential of the learned dynamics using the decomposition method \cite{lin2023computing}, of which details are present in \ref{app:id}. Subsequently, we plot the drift functions and the generalized potential of the learned and the exact dynamics in \cref{fig:2d}. Here, the reference solution is derived by numerically solving the FPK equation using the finite difference method. The results demonstrate agreements between both solutions when the step size $\Delta t$ is small. However, the Euler-Maruyama approach exhibits limitations in learning accurate dynamics and generalized potentials for larger step sizes. In contrast, DynGMA consistently provides accurate results, as evidenced in the third column of \cref{fig:2d}.

Subsequently, we uniformly sample $2\times 10^3$ initial points from the domain $[-2, 2] \times [-3, 3]$. Next, in the top portion of \cref{fig:2d_traj}, we present the distribution of these initial states and their subsequent evolution, following both the exact dynamics and the dynamics learned by DynGMA when $\Delta t = 0.2$. In the bottom portion of \cref{fig:2d_traj}, we display the mean and standard deviation of both the exact trajectories and learned trajectories for $\Delta t = 0.05, 0.1, 0.2$. The results demonstrate that the SDE learned using DynGMA more accurately captures the true dynamics.

In addition, we compute $e_{\sigma}$, $e_f$, and $e_V$ at gird points in $[-2,2]\times[-3,3]$, and record the errors in \cref{tab:error2d}, to quantitatively assess accuracy. The results indicate that DynGMA achieves lower errors when the step size $\Delta t$ is small and exhibits robust performance, as an increase in step size does not significantly impact its accuracy.
These findings illustrate that DynGMA is capable of learning more accurate dynamics, consequently resulting in a more precise invariant distribution, particularly when dealing with moderately large time steps.

\begin{figure}[!ht]
\centerline{\includegraphics[width=1\linewidth]{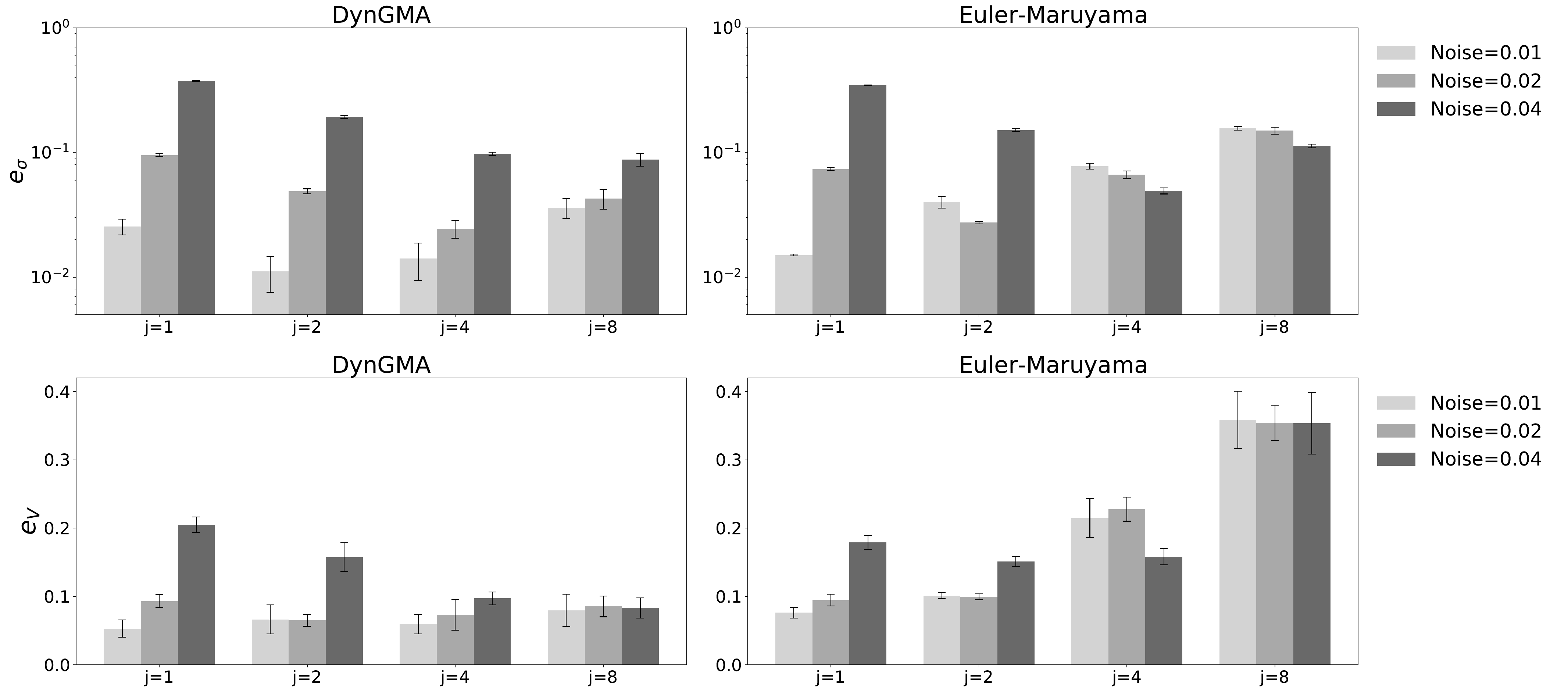}}
\caption{Errors in the learned diffusion coefficients and the generalized potential of the learned dynamics for multi-step loss} when using data with measurement noise for the two-dimensional system (\ref{eq:2d}). The results are obtained by taking the mean of $5$ independent experiments, and the black error bar represents one standard deviation.
    \label{fig:2d noise}
\end{figure}
Next, we show that multi-step training using DynGMA improves robustness in the presence of measurement noise in the system observations. We generate $4000$ trajectories and sample $41$ states at equidistant time steps with $\Delta t =0.025$. Each trajectory point is multiplied by $1+\delta$ to generate the training dataset, where $\delta$ follows a uniform distribution on $[-\text{Noise}, \text{Noise}]$, and the intensity $\text{Noise}$ is set to $0.01$, $0.02$, and $0.04$. We optimize the multi-step loss (\ref{eq:mloss}), where $\Gamma=\{j\}$ and $j$ takes values $1, 2, 4, 8$. As the Euler-Maruyama approach is only valid for one step, we directly expand its step size, take 
\begin{equation}\label{eq:multi-EM}
p_{h}(\Delta t_{m,\gamma}, y^{n}_{m\gamma}|0,  y^{n}_{(m-1)\gamma}; \theta) = \N(y^{n}_{m\gamma}|y^{n}_{(m-1)\gamma} + \Delta  t_{m,\gamma} f_{\theta}(y^{n}_{(m-1)\gamma}), \Delta  t_{m,\gamma}\sigma_{\theta}(y^{n}_{(m-1)\gamma}) \sigma^{\top}_{\theta}(y^{n}_{(m-1)\gamma})),
\end{equation}
and substitute it into loss (\ref{eq:mloss}).
Note that excessive measurement noise makes it challenging to distinguish between measurement noise and the true randomness induced by diffusion. Thus, we record the error of the learned diffusion and the learned potential in \cref{fig:2d noise}. It depicts that the error increases as the noise intensity increases when employing small steps ($j=1,2$) for both methods. With multi-step training involving more steps, the discrepancy in error levels among the results obtained using DynGMA under different noise intensities diminishes. However, augmenting the number of steps in the Euler-Maruyama approach results in larger time step sizes, consequently resulting in notable numerical errors. In contrast, DynGMA is capable of handling data with larger step sizes, allowing it to deal with measurement noise by combining multi-step training.

\begin{figure}[!ht]
\centerline{\includegraphics[width=1\linewidth]{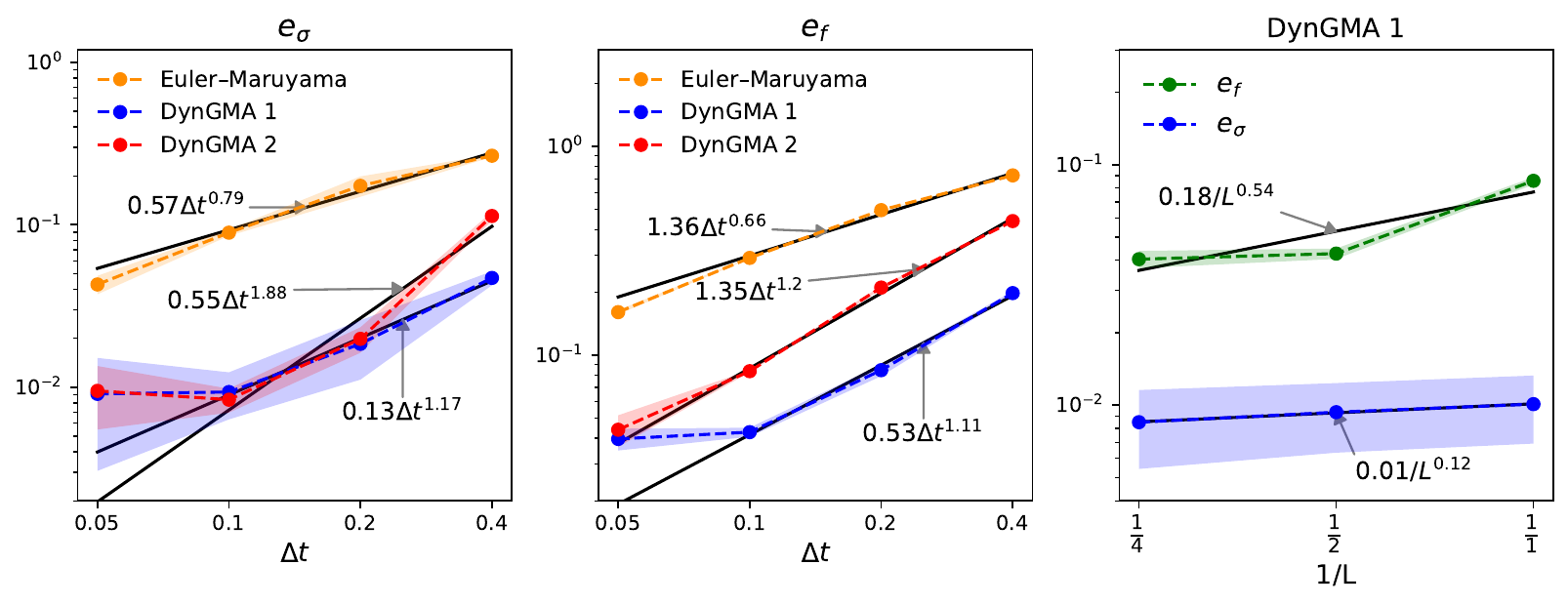}}
\caption{(Left and Middle) convergence of error of DynGMA and the Euler-Maruyama approach with respect to time step size $\Delta t$. (Right) convergence of error of DynGMA with respect to parameter $L$.}
    \label{fig:2d_order}
\end{figure}
Finally, to numerically verify the accuracy of the Gaussian density approximation demonstrated in Theorem \ref{the:dis con}, \ref{the:dis con2} and \ref{the:dis con3}, we test the convergence of the errors incurred by Algorithm 1 (with $K=1$ and $L=2$), Algorithm 2 (with $K=1$), and the Euler-Maruyama approach across varying data step sizes ($\Delta t=0.05, 0.1, 0.2, 0.4$), while retaining the same data generation parameters and network settings. The left and middle panels of \cref{fig:2d_order} illustrate the convergence of errors for DynGMA and the Euler-Maruyama approach with respect to $\Delta t$. It is observed that the errors of all methods decrease as $\Delta t$ decreases. Eventually, both DynGMA methods reach a plateau where the rate of error reduction levels off. In addition to the discretization error induced by density approximation, the learning errors of neural networks—encompassing generalization, optimization, and approximation errors—can also influence the final outcome. Therefore, to calculate the convergence order, we construct linear regression using only the errors for $\Delta t=0.1, 0.2, 0.4$ where numerical error accounts for the dominant term. The black solid line here represents the result of regression. DynGMA with Algorithms 1 and 2 consistently demonstrate a higher convergence order. The right panel of \cref{fig:2d_order} shows the convergence of error for DynGMA using Algorithm 1 with respect to parameter $L$. Increasing $L$ from $1$ to $2$ results in a notable decrease in error. However, further increments in $L$ do not yield significant improvement as learning error may dominated. As our theoretical analysis focuses on the error of the density approximation, it is not able to directly quantify the error between the true governing function and its approximation. We would like to further investigate this problem in future research.

\subsection{Learning SDEs with state-dependent diffusions}\label{ssec:ls}

Subsequently, we consider the $3$d Lorenz system \cite{lorenz1963deterministic,sparrow2012lorenz}, which is a simplified model for atmospheric dynamics and is widely investigated for learning dynamics \cite{lin2023computing, yu2021onsagernet}. To showcase the effectiveness of the proposed method in learning SDEs with state-dependent diffusions, we investigate the Lorenz system in the presence of multiplicative noise \cite{chekroun2011stochastic}, formulated as: 
\begin{equation}\label{eq:LS}
\begin{aligned}
\d x &= \brac{-\sigma x - \sigma y}\d t &&+ \varepsilon x \d \omega_1,\\
\d y &= \brac{-xz + rx -y} \d t &&+ \varepsilon y \d \omega_2,\\
\d z &= \brac{xy-bz}\d t  &&+ \varepsilon z \d \omega_3,
\end{aligned}
\end{equation}
where the geometric parameter $b =8/3$, the Prandtl number $\sigma=10$, the rescaled Rayleigh number $r=28$, and the noise intensities are set as $\varepsilon=0.3$. Although its diffusion matrix is not strict positive definite, the following results indicate that our algorithm remains effective. 

We simulate $N=10^4$ trajectories for the training dataset. For each trajectory, the initial state is sampled from the uniform distribution on the domain $[-25, 25]\times [-30, 30]\times [-10, 60]$, and we collect $M=200$ pairs of data points at the time slices $\{t_j= 0.01*j: \ j =0, 1, \cdots, M-1\}$. In total, there are $2\times 10^6$ pairs of points. Here, \cref{alg:dyngma2} with $h=0.005$, corresponding to $K=2$, is used to obtain the approximation of the transition density. And the results are compared with those obtained using the Euler-Maruyama and Gaussian Cubature methods (employing the Euler method with a step size of $0.005$).
In all methods, the drift $f_{\theta}$ is parameterized in the same manner as described in \cref{ssec:2d}, and the diffusion $\sigma_{\theta}$ is parameterized according to (\ref{eq:sigma}), where the used FNN has one hidden layer with $50$ units and tanh activations. The optimization process employs the Adam optimization algorithm, with a chosen batch size of $10^5$. The learning rate is set to $10^{-3}$ for DynGMA and the Euler-Maruyama and $10^{-4}$ for the Gaussian Cubature approach. Results are collected after $10^5$ parameter updates.

\begin{table}[ht]
  \centering
    \resizebox{0.6\linewidth}{!}{
    \begin{tabular}{c ccc }
    \toprule 
    {Methods}& DynGMA&Euler-Maruyama& Gaussian Cubature \\
    \midrule
    $e_f$&7.83e-2$\pm$9.29e-3 & 1.27e-1$\pm$3.52e-3 & 1.65e-1$\pm$1.15e-2 \\
    $e_{\sigma}$&1.54e-1$\pm$4.80e-2 & 1.96e-1$\pm$1.16e-2 & 5.82e-1$\pm$2.18e-1 
    \\
    $e_p$ & 6.36e-2$\pm$5.33e-3 & 6.38e-1$\pm$1.74e-2 & 2.45e-1$\pm$1.37e-2 \cr
    \bottomrule
    \end{tabular} 
    }
    \caption{Quantitative results of the Lorenz system (\ref{eq:LS}). The results are obtained by taking the mean of $5$ independent experiments, and the errors are recorded in the form of mean $\pm$ standard deviation.}
    \label{tab:errorls}
\end{table}
\begin{figure}[htbp]
\centerline{\includegraphics[width=1\linewidth]{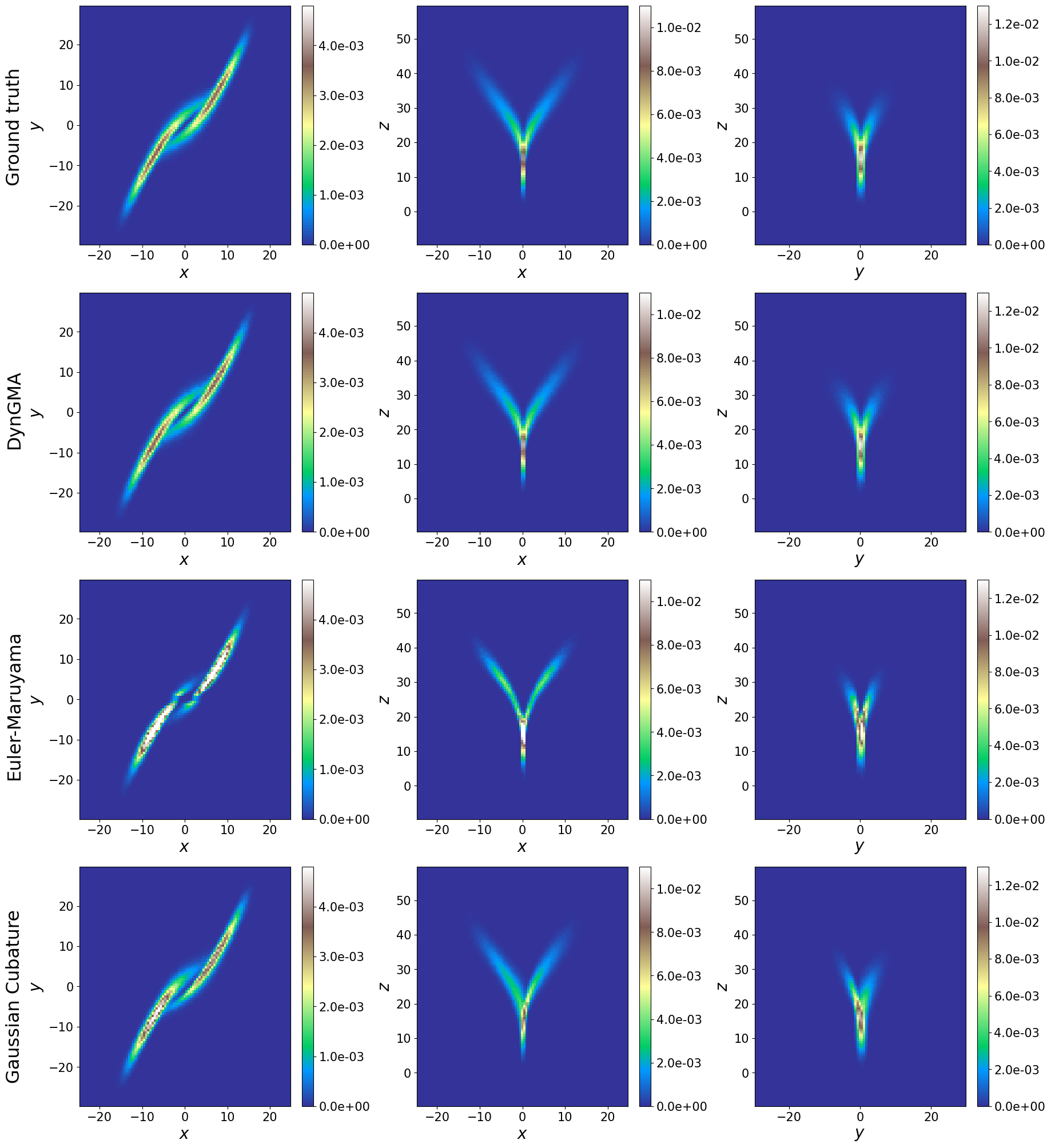}}
\caption{Invariant distribution of true dynamics and learned dynamics. The two distributions are projected onto the three planes with $z=25$(left), $y =0$(middle) and $x =0$(right), respectively.
}
    \label{fig:ls}
\end{figure}

After training, we estimate the invariant distribution using the Monte Carlo method to further validate the learned dynamics. In \cref{fig:ls}, we present a comparison of three cross-sections of the invariant distribution of the true dynamics and that of the learned dynamics using various methods. It is difficult to distinguish between the invariant distribution of dynamics learned by DynGMA and that of exact dynamics. In contrast, the invariant distribution of dynamics learned by the Euler-Maruyama and Gaussian Cubature approaches displays notable differences, particularly in attractive basins with high probabilities.

Moreover, we compute the errors $e_f$, $e_{\sigma}$, and $e_p$, evaluated at $10^5$ points uniformly sampled in $[-25, 25]\times [-30, 30]\times [-10, 60]$. The results are presented in \cref{tab:errorls} for various methods. It is shown that our proposed DynGMA outperforms other methods in terms of accuracy. Interestingly, despite the Gaussian Cubature having a discrete step size smaller than that of the Euler-Maruyama approach, it fails to yield a substantial improvement. It is noted that the Gaussian Cubature method involves two steps, and an excessive number of steps can potentially lead to non-positive covariance, resulting in the failure of the training process. Currently, the application of the Gaussian Cubature method in machine learning is limited to time series prediction using SDE as black-box \cite{solin2021scalable}. Its utilization for learning interpretable drift and diffusion functions in SDE models has not been explored. Further research might be required to investigate the parameterization and discretization of this method for such tasks.

\subsection{Learning high-dimensional SDEs}\label{ssec:emt}
We next validate the performance of the DynGMA in learning a $10$d biological model combined by the epithelial-mesenchymal transition (EMT) network and the metastasis network. Researchers can investigate the correlation between EMT and the metastasis of cancer via this biological model \cite{heerboth2015emt} and have proposed several works to compute its invariant distribution \cite{li2018landscape, lin2023computing}. The Euler-Maruyama loss has been employed to learn its stochastic dynamics and invariant distributions from data \cite{lin2023computing}. Herein, we use a similar experimental setup but a larger time step size.

Specifically, the training data is obtained by solving the governing SDE with constant diffusion, of which the governing function is given in \ref{app:emt}. We first sample $10^5$ initial states from the uniform distribution on the domain $\Omega_1\times \cdots \times \Omega_{10}$ where $\Omega_i=[0, 2]$ for $i \neq 7$ and $\Omega_7=[0, 6]$, then simulate trajectories and collect $36$ pairs of data points at the times $\{t_j, t_j+\Delta t\}$, where $t_j =2, 2.5,\cdots, 19.5$. In total, there are $7.2 \times10^6$ data points. The step size $\Delta t$ is set as $0.005$ in previous work \cite{lin2023computing} and here we configure $\Delta t$ as $0.04$ or $0.08$. Due to considerations of computational efficiency and resources, we apply the Adam optimization with a batch size of $10^5$ for $10^5$ iterations to train the model, and we use \cref{alg:dyngma1} with $K=1$ and $L=2$, i.e., Gaussian density approximation with (\ref{eq:ite of con}). It is noted that in high-dimensional scenarios, DynGMA with larger $K$ would significantly increase computational complexity. Therefore, herein we solely employ the proposed Gaussian approximation. The chosen model architecture and other hyperparameters remain consistent with those outlined in \cref{ssec:2d}.

\begin{figure}[htbp]
\centerline{\includegraphics[width=1\linewidth]{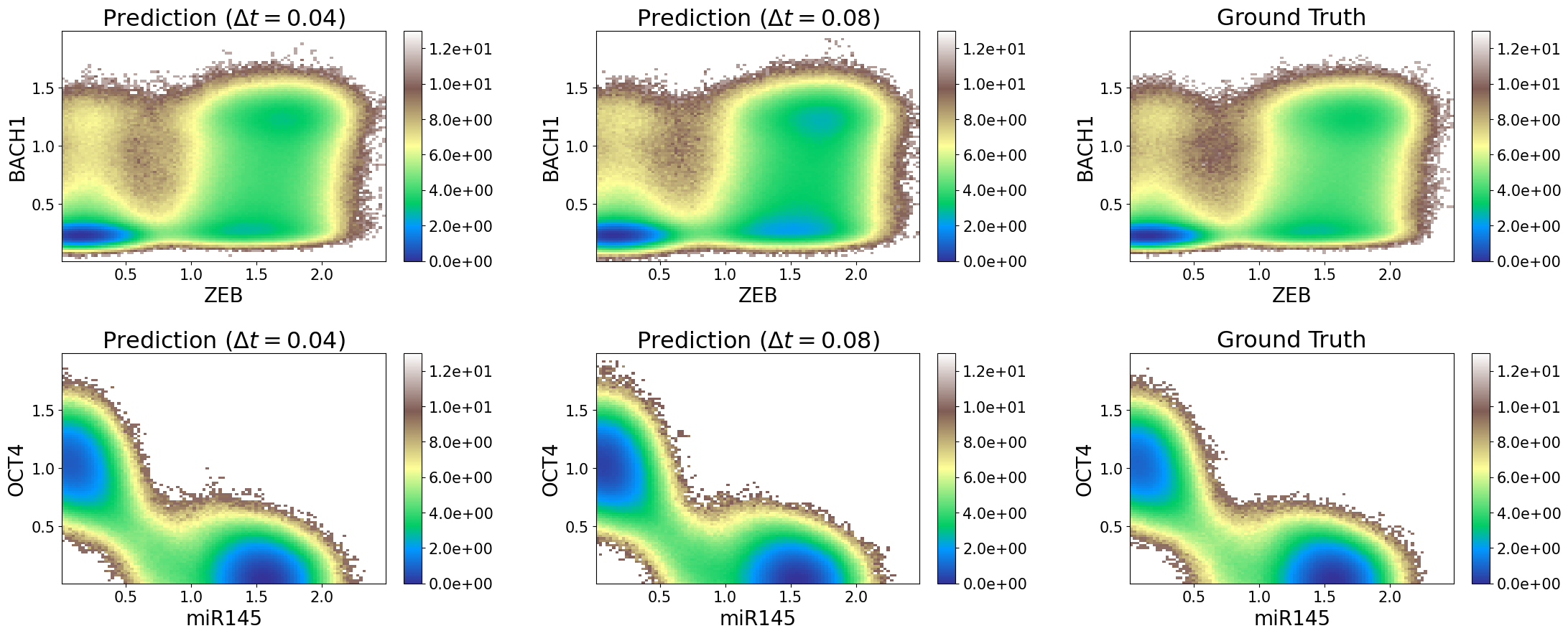}}
\caption{Free energy of learned dynamics and exact dynamics for the EMT-metastasis network.
}
    \label{fig:emt}
\end{figure}
 
After training, we compute the free energy to assess the accuracy of the learned dynamics, which is given by
\begin{equation*}
V_{2,10}(x_2, x_{10})= -\log \int_{\R^8} p(x_1,\cdots, x_{10}) \d x_1 \d x_3 \cdots \d x_9,
\end{equation*}
where $p(x_1,\cdots, x_{10})$ is the invariant distribution of the dynamical system. Since the governing equation has been learned, we simulate long trajectories of the learned SDE to sample the invariant distributions and then estimate the marginal distribution of the two variables. We also compute the free energy landscapes $V_{6,3}$ for another two variables, $x_6$ and $x_3$, in the same manner. The resulting free energies are depicted in \cref{fig:emt}. We can observe a strong agreement between the results obtained from the learned dynamics and the exact dynamics.
In addition, we evaluate the relative errors of the learned diffusion matrix. When employing our proposed method, the relative errors are $e_{\sigma}=0.0798$ ($\Delta t=0.04$) and $e_{\sigma}=0.0998$ ($\Delta t=0.08$). In comparison, training the network using the Euler-Maruyama loss leads to higher relative errors for the learned diffusion matrix: $e_{\sigma}=0.1882$ ($\Delta t=0.04$) and $e_{\sigma}=0.2354$ ($\Delta t=0.08$). These results demonstrate that our approach remains effective for moderately high-dimensional equation discovery and invariant distribution computation.

\subsection{Learning effective SDEs from Gillespie's stochastic simulations data with variable time step-sizes}\label{ssec:sir}
In the subsequent sections, we validate the effectiveness of the proposed DynGMA in handling data with variable, and even uncontrollable, time step-sizes. We consider learning effective SDEs from trajectories generated by Gillespie's stochastic simulation algorithm (SSA), where an exact SDE does not exist.
SSA generates sample sequences of continuous-time Markov chains, where individual events occur with random timing. 
The time steps between reaction events follow an exponential distribution, resulting in variable and uncontrollable step sizes. 
Specifically, We consider the Susceptible-Infectious-Recovered (SIR) model, an epidemiological compartment model used to simulate the spread of infectious diseases in a population. In this model, individuals can transition between three states: Susceptible (S), Infectious (I), and Recovered (R). For a total population size of $N$, we denote the counts of susceptible, infected, and recovered individuals as $n_0$, $n_1$, and $n_2$, respectively. Their corresponding concentrations are represented by $y_0$, $y_1$, and $y_2$. At each time step of Gillespie's SSA, the rates $r_1 = 4k_1y_0y_1$, $r_2 = k_2y_1$, and $r_3 = k_3y_2$ are computed, and subsequently, $y_1$ and $y_2$ are updated accordingly. The mass balance then determines that the concentration of susceptible individuals is given by $y_0=1-y_1-y_2$.

\begin{figure}[htbp]
\centerline{\includegraphics[width=1\linewidth]{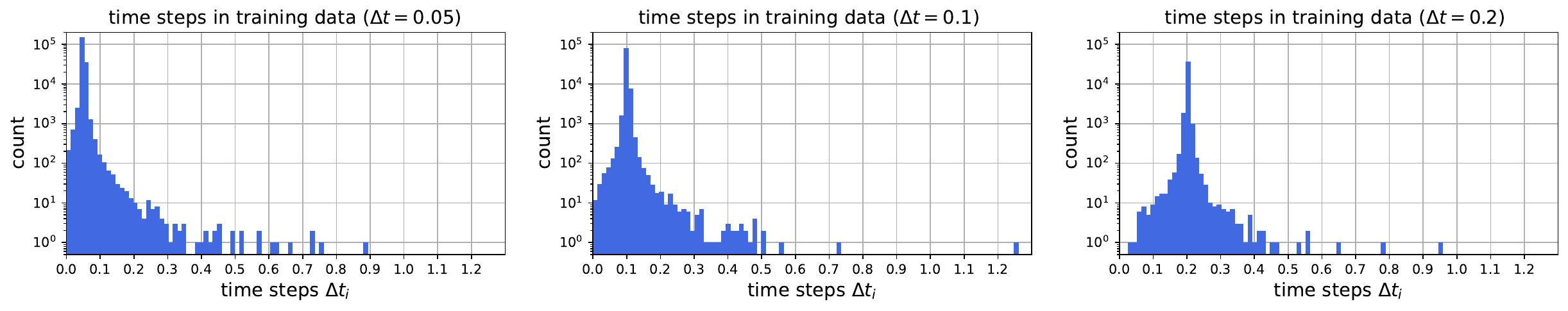}}
\caption{Time steps in training data for various settings of $\Delta t$.
}
    \label{fig:kmch}
\end{figure}
\begin{figure}[ht]
\centerline{\includegraphics[width=1\linewidth]{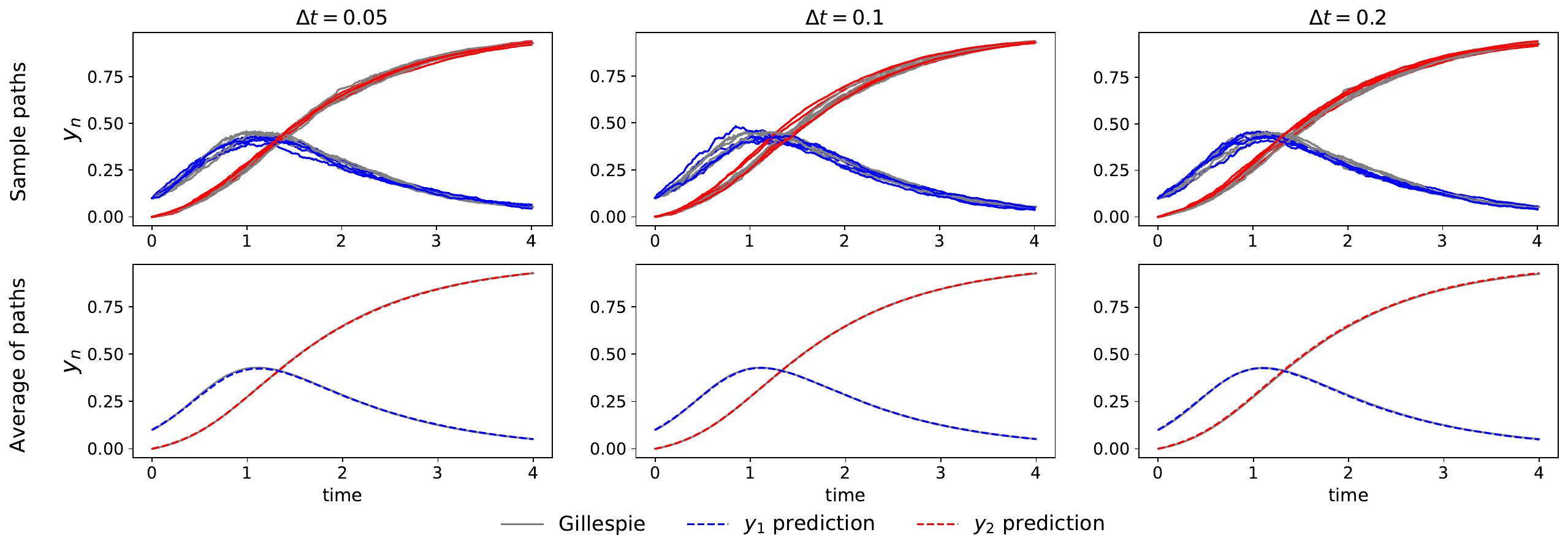}}
\caption{$5$ sample paths and average of $2000$ paths of the learned SDE using DynGMA for SIR model.
}
    \label{fig:kmc}
\end{figure}
\begin{figure}[!ht]
\centerline{\includegraphics[width=1\linewidth]{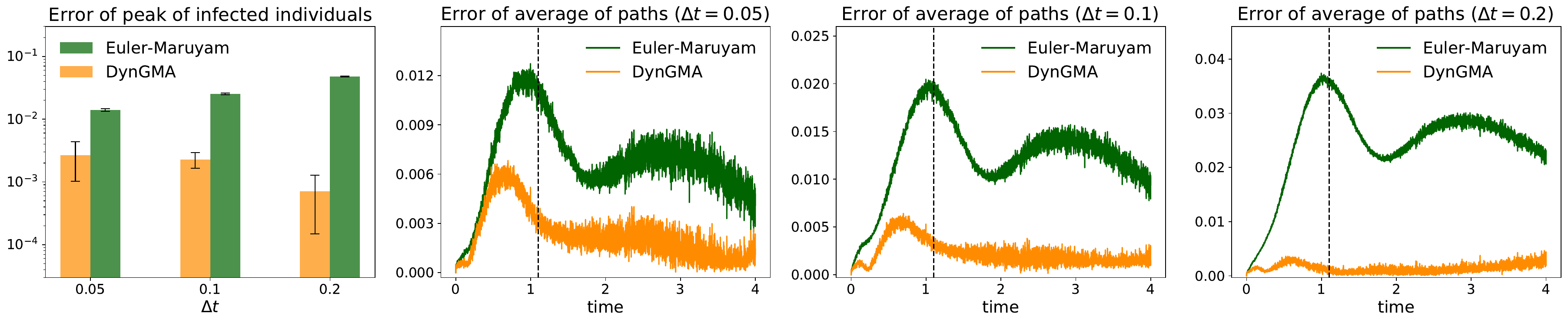}}
\caption{Error of the learned statistics of $2000$ paths. (Left) The results are obtained by taking the mean of $5$
independent experiments, and the black error bar represents one standard deviation.  (Three columns on the right) The black dashed line is the moment when the exact number of infected individuals reaches its maximum value.
}
    \label{fig:kmcerror}
\end{figure} 

In Gillespie's SSA, the parameters $t_{max}$ and $\Delta t$ are predetermined, and the data points are evaluated at ${t_0, t_1, \cdots, t_{max}}$, where $\Delta t_i = t_i-t_{i-1} \approx \Delta t$. In this experiment, the model parameter settings for the SIR model are as follows:  $k_1=1$, $k_2=1$, $k_3=0$ and $N=1024$. We set $t_{max}=1$ and choose various values for $\Delta t$ to generate $10^5$ trajectories for each training dataset and $2.5\times 10^4$ trajectories for each test dataset. The data step size is a random variable and is illustrated in \cref{fig:kmch} for different settings of $\Delta t$. There exist some uncontrollably large steps $\Delta t_i$ for all settings of $\Delta t$. We use one hidden layer in FNN with $64$ units and tanh activations to represent drift $f_{\theta}$, and parameterize diffusion $\sigma_{\theta}$ according to (\ref{eq:sigma}). Adam optimization with the learning rate of $10^{-3}$ is used to update parameters for $2 \times 10^4$ epochs. Under these settings, we carry out \cref{alg:dyngma2} (if $\Delta t=0.2$, then $K=2$; otherwise, $K=1$) to obtain the approximation and optimize the loss function (\ref{eq:loss}) to train the networks.

Since there is no exact SDE for this problem, we perform predictions starting from initial conditions $y_1=0.1$, $y_2=0$ to measure the learning performance. As depicted in \cref{fig:kmc}, the network trained using DynGMA accurately captures the stochastic dynamics of the results from SSA simulations. Subsequently, we use the Euler-Maruyama approach to train networks under the same settings for comparison. We estimate the average peak of infected individuals across $2000$ paths for both the learned and true dynamics, recording the errors in the first column of \cref{fig:kmcerror}. In this analysis, we conduct $5$ independent experiments to obtain the means and standard deviations. Additionally, we plot the error of the average paths in the three columns on the left side of \cref{fig:kmcerror}. Notably, DynGMA achieved the lowest error, and as $\Delta t $ increased, its improvement became more significant. It is noted that the error of the Euler-Maruyama approach reaches its maximum value almost synchronously with the peak of infected individuals, which corresponds to the time of the most intense reaction. This observation indicates that the error of the Euler-Maruyama approach is primarily influenced by numerical errors.

\begin{figure}[htbp]
\centerline{\includegraphics[width=1\linewidth]{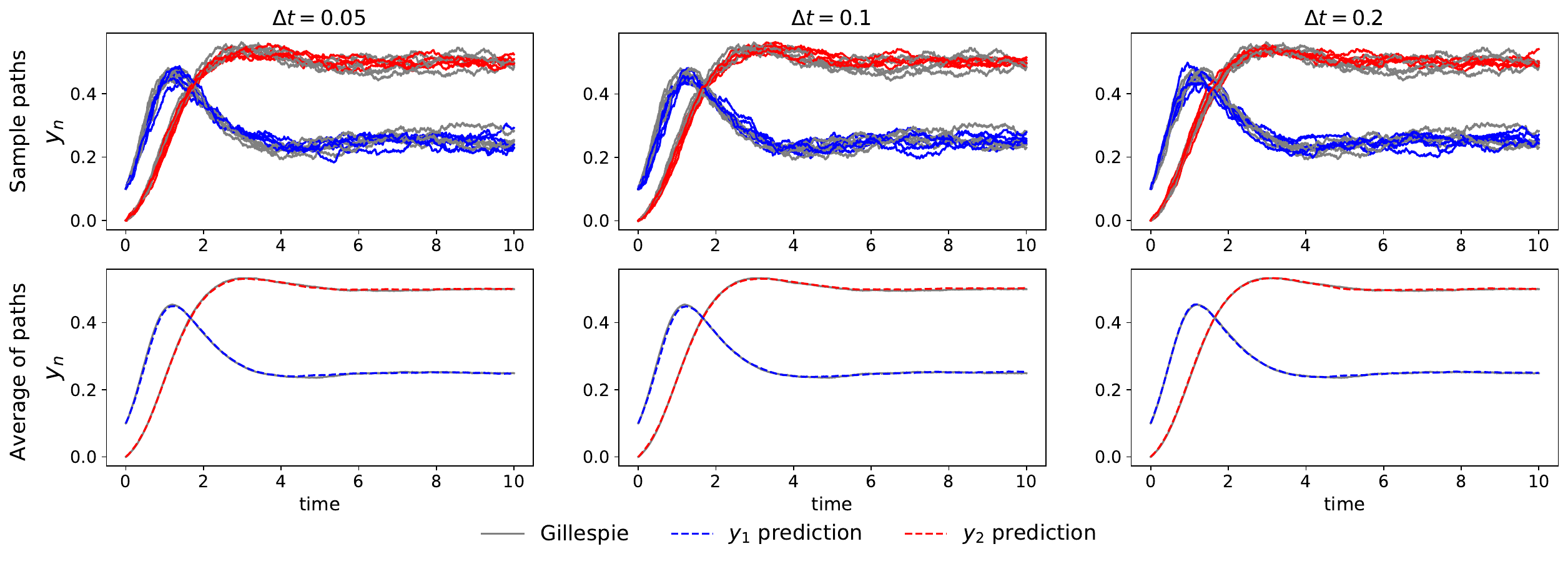}}
\caption{$5$ sample paths and the average of $2000$ paths of the learned SDE using DynGMA for SIRS model.
}
    \label{fig:kmc_sirs}
\end{figure}

Next, we consider the SIRS model by setting $k_3=0.5$, where recovered individuals have a probability of being reinfected and richer dynamic behaviors are exhibited. We generate the data for the SIRS model in the same manner and use the same settings to train networks. The predictions are shown in \cref{fig:kmc_sirs}. It is shown that the learned network model captures stochastic dynamics perfectly and is capable of reconstructing the correct oscillations and steady state.

\section{Summary}\label{sec:Conclusions}
In this paper, we present a novel method for approximating the transition density of a Stochastic Differential Equation (SDE) aimed at learning unknown stochastic dynamics from data. The proposed density approximation allows for multiple time steps, thereby obtaining a more accurate approximation for large time interval, while existing methods are usually based on one-step stochastic numerical schemes and thus necessitate data with sufficiently high time resolution. Our approach includes a Gaussian density approximation inspired directly by the random perturbation theory of dynamical systems \cite{blagoveshchenskii1962diffusion, blagoveshchenskii1961certain}, complemented by an extension named as dynamical Gaussian mixture approximation (DynGMA). The targeted SDE describes dynamical systems with random perturbations. While its drift and diffusion functions are fully unknown, discrete observations of certain trajectories are available.
To reconstruct unknown stochastic dynamics, we leverage neural networks to approximate both the drift and diffusion functions of the SDE. Subsequently, we employ the proposed density approximation to construct a likelihood loss, which serves as the training objective for the neural networks. 
The proposed method is capable of reconstructing unknown governing functions of SDEs through a data-driven approach, and thus can be applied to invariant distribution computing from data.
Benefiting from the robust density approximation, our method demonstrates robustness and possesses many benefits across various application scenarios. 
In particular, it exhibits superior accuracy with a given step size, remains effective even when confronted with trajectory data with low time resolution, and performs well even in the presence of higher levels of measurement noise. 
Finally, we validate the superior properties of our proposed method in comparison to competing baseline models in a variety of examples across various scenarios, including a biological network model and an epidemiological compartment model.

There are many possible extensions of our work.
As we have provided the error quantification of the Gaussian approximations, it is interesting to further explore the interplay between the training process and numerical integration to adapt the time step. A possible direction is to estimate the magnitude of the current diffusion and adaptively select the discretization accuracy based on it in the training process. Additionally, exploring the combination of DynGMA with kernel methods \cite{hamzi2023note}, as well as extending DynGMA to forward solving SDEs \cite{li2021numerical}, could be interesting directions. In terms of theory, the quantification of the error between the learned and target governing functions, as well as the analysis of convergence in subsequent invariant distribution computation, remains an open challenge. It is important to note that the error of the approximated transition density does not directly yield the aforementioned bounds. While there have been previous analyses on this task \cite{gu2023stationary}, their analysis focuses on SDE with constant diffusion matrix, and the loss function is derived through the one-step Euler-Maruyama scheme.
We would like to explore the feasibility of extending existing analysis results to our method in future research. 
In addition, DynGMA and its application in computing invariant distributions from data are specifically designed for unknown SDEs driven by Gaussian noise. We would like to derive methods capable of effectively handling general non-Gaussian noise in future research.
It is worth mentioning that DynGMA operates independently of specific structures, and the $f_{\theta}$ and $\sigma_{\theta}$ can be any valid neural networks or functions, depending on the specific problem. 
Our algorithm naturally accommodates the encoding of partially known physical terms, such as generalized Onsager principle \cite{chen2023constructing, yu2021onsagernet}, symplectic structure \cite{bertalan2019learning, greydanus2019hamiltonian}, and GENERIC formalism \cite{zhang2022gfinns}. Using our density approximation to learn structure-preserving stochastic dynamics might be an interesting direction.
Moreover, in practical applications, many high-dimensional systems necessitate order reduction, requiring simultaneous learning of the latent states and the governing function. An example includes the learning of the macroscopic thermodynamic description of the stretching dynamics of polymer chains \cite{chen2023constructing}. The investigation of the performance of our method in handling such tasks is also left for future work.

\section*{Acknowledgments}
This project is supported by the National Research Foundation, Singapore, under its AI Singapore Programme (AISG Award No: AISG3-RP-2022-028) and the NRF fellowship (project No. NRF-NRFF13-2021-0005).

\appendix
\allowdisplaybreaks
\section{EMT-metastasis network}\label{app:emt}
\renewcommand{\vec}[1]{\boldsymbol{#1}}
In the EMT-metastasis network, the governing SDE is expressed as follows:
\begin{equation*}
\d \vec{y} = \vec{f}(\vec{y}) + 0.2 \, \d \vec{\omega}.
\end{equation*}
Here, $\vec{y} = (y_1, \cdots, y_{10})$ represents the relative expression levels of ten genes in the network, and $\vec{\omega}$ is a ten-dimensional standard Brownian motion. The drift function $\vec{f}=(f_1, \cdots, f_{10})$ is defined by: 
\begin{align*}
f_1 &= 0.8  \brac{ \frac{0.5}{0.5+y_1} + \frac{0.5^2}{0.5^2+y_5^2} + \frac{0.5^4}{0.5^4+y_7^4} } - y_1, \\     
f_2 &= 0.2 \left( \frac{y_1}{0.5 + y_1} + \frac{y_2^2}{0.5^2 + y_2^2} \right) + 0.8 \left( \frac{0.5^6}{0.5^6 + y_4^6} + \frac{0.5^4}{0.5^4 + y_6^4} \right) - y_2,  \\
f_3 &= 0.2 \left( \frac{y_3^2}{0.5^2 + y_3^2} + \frac{y_9^4}{0.5^4 + y_9^4} \right) + 0.8 \frac{0.5^4}{0.5^4 + y_6^4} - y_3,  \\
f_4 &= 0.2 \frac{y_3^4}{0.5^4 + y_3^4} + 0.8 \left( \frac{0.5}{0.5 + y_1} + \frac{0.5^3}{0.5^3 + y_2^3} \right) - y_4,  \\
f_5 &= 0.8 \left( \frac{0.5}{0.5 + y_1} + \frac{0.5^2}{0.5^2 + y_2^2} \right) - y_5,  \\
f_6 &= 0.8 \left( \frac{0.5^4}{0.5^4 + y_2^4} + \frac{0.5^4}{0.5^4 + y_3^4} \right) - y_6,  \\
f_7 &= 0.8 \frac{y_7^2}{0.5^2 + y_7^2} + 7 \frac{y_8^5}{0.5^5 + y_8^5} + 0.8 \frac{0.5^4}{0.5^4 + y_9^4} - 4 y_7 y_{10} - y_7,  \\
f_8 &= 0.8 \left( \frac{0.5^4}{0.5^4 + y_1^4} + \frac{0.5}{0.5 + y_{10}} \right) - y_8 , \\
f_9 &= 0.2 \frac{y_9^2}{0.5^2 + y_9^2} + 0.8 \frac{0.5^4}{0.5^4 + y_7^4} - y_9,  \\
f_{10} &= 0.1 + \frac{4}{1 + y_{10}^3} - 4 y_7 y_{10} - y_{10}.
\end{align*}

\section{Computing invariant distributions from data}\label{app:id}
Our learning model is able to discover the dynamics and thus can be turned into an invariant distribution computation algorithm for the situation where we have access to noisy data but the governing equation is unknown. We next introduce the detailed algorithm following \cite{lin2023computing}. 

The density function $p(x) $ of the invariant distribution of the SDE (\ref{eq:sde}) satisfies the stationary FPK equation:
\begin{equation}\label{eq:id_p}
- \sum_{d=1}^D \frac{\partial }{\partial x_d} \{ [f(x)]_d p(x) \} + \sum_{d=1}^D \sum_{d'=1}^D \frac{\partial^2 }{\partial x_{d_1}\partial x_{d'}} \{ [\sigma(x)\sigma^{\top}(x)]_{dd'} p(x) \} = 0,\quad \int_{\R^D} p(x) \d x=1.
\end{equation}
Following \cite{lin2023computing}, we define the generalized potential
\begin{equation*}
V(x) = -\varepsilon \log p(x).
\end{equation*}
Then, by writing the diffusion function 
as $\sigma(x) = \sqrt{2\varepsilon}\bar{\sigma}(x)$ where $\snorm{\bar{\sigma}}_F=1$ and $\bar{\Sigma} := \bar{\sigma}\bar{\sigma}^{\top}$, $V(x)$ satisfies
\begin{equation}\label{eq:id_V0}
(f(x)+ \bar{\Sigma}(x) \nabla V(x) - \varepsilon \d\bar{\Sigma}(x)) \nabla V(x) -\varepsilon \nabla \cdot (f(x) + \bar{\Sigma}(x) \nabla V(x) - \varepsilon \d\bar{\Sigma}(x)) = 0,
\end{equation}
where $\d\bar{\Sigma}(x) \in \R^D$ and $(\d\bar{\Sigma}(x))_d = \sum_{d'=1}^D \frac{\partial}{\partial x_{d'}} \bar{\Sigma}_{dd'}(x)$. Equivalently, equation (\ref{eq:id_V0}) can be written as:
\begin{subequations}  \label{eq:id_V}
\begin{align}
-\bar{\Sigma}(x) \nabla V(x) + \varepsilon \d\bar{\Sigma}(x) + g(x) &=f(x), \label{eq:id_V1} \\
g(x) \nabla V(x) -\varepsilon \nabla \cdot g(x) &= 0.\label{eq:id_V2} 
\end{align}
\end{subequations}
Focusing on constant diffusion (i.e., $\d\bar{\Sigma}\equiv 0$), \cite{lin2023computing} proposes to identify the drift and diffusivity by parameterizing $f$ via (\ref{eq:id_V1}) and employing the Euler–Maruyama method to approximation transition density, and simultaneously compute the invariant distribution by network-based residual minimization according to (\ref{eq:id_V2}). Alternatively, we first identify a approximation of $f$ and $\bar{\Sigma}$, i.e., $f_{\theta^*}$ and $\sigma_{\theta^*}$ by optimizing (\ref{loss}) or (\ref{multi-steploss}) and then employ network-based residual minimization for solving (\ref{eq:id_V}). Specifically, we parameterize the generalized potential $V$ and the auxiliary function $g$ as $V_{\phi}$ and $g_{\phi}$ respectively. The parameters $\phi$ can be obtained by minimizing the physical-informed loss, i.e.,
\begin{equation*}
\begin{aligned}
\phi^* = \arg \min_{\phi} \sum_{n=1}^N\sum_{m=0}^M& \norm{-\bar{\Sigma}_{\theta^*}(y^n_m) \nabla V_{\phi}(y^n_m)  + g_{\phi}(y^n_m) -f_{\theta^*}(y^n_m)}_2^2  + \norm{g_{\phi}(y^n_m)  \nabla V_{\phi}(y^n_m) -\varepsilon_{\theta^*} \nabla \cdot g_{\phi}(y^n_m) }_2^2,\\
\end{aligned}
\end{equation*}
where $\varepsilon_{\theta^*} = \norm{\sigma_{\theta^*}}_2^2/2$, $\bar{\Sigma}_{\theta^*} =\sigma_{\theta^*}\sigma_{\theta^*}^{\top}/2/\varepsilon_{\theta^*}$, and $y^n_m$ are the observations of given $N$ trajectories. Adam optimizer \cite{kinga2015method} is applied to update the parameters. After training, we shift the potentials $V_{\phi^*}$ so that their minimum value in the domain is equal to zero. 

{
\section{Learning Bene\v{s} SDE from data}\label{app:Learning Benes SDE}
Herein, we investigate the performance of DynGMA in learning Bene\v{s} SDE from data. As we focus on dynamical systems with random perturbations, we reduce the intensity of diffusion, considering
\begin{equation*}
\d x = \tanh x \d t + 0.3 \d \omega.
\end{equation*}
We sample $5\times 10^5$ points from the uniform distribution on $[-1,1]$ as initial values and compute the states after time step $\Delta t$ to obtain the training dataset. We employ a feedforward neural network (FNN) with one hidden layer containing 64 units and sigmoid activation functions to parameterize the drift $f_{\theta}$. Additionally, we use a trainable constant to parameterize the diffusion $\sigma_{\theta}$. We use Algorithm 1 to approximate the transition density, setting the step size of each sub-Gaussian step to $h = 0.2$ and the step size of discretization (\ref{eq:ite of con}) to $0.1$. This configuration corresponds to $K = \lceil \Delta t/0.2\rceil$ and $L = \min(0.1, \Delta t)/0.1$. Specifically, for $\Delta t$ values of $0.1$, $0.2$, and $0.4$, the values of $K$ are $1, 1, 2$, and the values of $L$ are $1, 2, 2$, respectively. During training, we employed full-batch Adam optimization for $5\times 10^3$ epochs, with the learning rate set to exponentially decay with linearly decreasing powers from $0.1$ to $0.01$.}

{
\begin{table}[ht]
  \centering
    \resizebox{\textwidth}{!}{
    \begin{tabular}{ccccccc}
    \toprule
    \multirow{2}{*}{Settings}&
    \multicolumn{2}{c}{$\Delta t=0.1$}&\multicolumn{2}{c}{$\Delta t=0.2$}&\multicolumn{2}{c}{$\Delta t=0.4$}\cr
    \cmidrule(lr){2-3} \cmidrule(lr){4-5} \cmidrule(lr){6-7}
    & DynGMA&Euler-Maruyama& DynGMA&Euler-Maruyama& DynGMA&Euler-Maruyama\cr
    \midrule 
$e_{\sigma}$
& 2.37e-3$\pm$1.00e-3 & 7.44e-2$\pm$2.14e-3 & 8.78e-4$\pm$9.24e-4 & 2.37e-1$\pm$1.63e-1 & 2.42e-2$\pm$4.06e-3 & 3.30e-1$\pm$4.97e-3 \cr
$e_f$
& 6.06e-3$\pm$2.03e-3 & 3.01e-2$\pm$2.43e-3 & 6.24e-3$\pm$1.37e-3 & 6.25e-2$\pm$1.08e-2 & 1.08e-2$\pm$3.32e-3 & 1.10e-1$\pm$1.16e-3 \cr
    \bottomrule
    \end{tabular} 
    }
    \caption{Quantitative results for learning Bene\v{s} SDE. The results are obtained by taking the mean of $5$ independent experiments, and the errors are recorded in the form of mean $\pm$ standard deviation.}
    \label{tab:error1d}
\end{table}
After the training process is completed, we calculate $e_{\sigma}$ and $e_f$ at $1000$ grid points within the interval $[-1,1]$, and document the errors in Table \ref{tab:error1d} to quantitatively evaluate accuracy. The findings demonstrate that DynGMA also achieves lower errors, consistent with the results presented in \cref{sec:Numerical experiments}.
}

\section{Proofs}\label{app:proofs}
\begin{proof}[Proof of \cref{cor:gauss}]
By rescaling time $\hat{t} = t/h$, and taking $\varepsilon=\sqrt{h}\norm{\sigma}_F$, the proof is a direct consequence of \cref{thm:gauss0}.
\end{proof}

\begin{proof}[Proof of \cref{the:dis con}]
We first show that the one-step scheme
\begin{equation*}
\begin{aligned}
&\Nmu_{l+1/2} = \Nmu(lh/L) + \frac{h}{2L} f_{\theta}(\Nmu(lh/L)), \quad
\Nmu_{l+1} = \Nmu(lh/L) + \frac{h}{L} f_{\theta}(\Nmu_{l+1/2});\\
&\NSigma_{l+1} = \left(\1_{D\times D} + \frac{h}{L}  J_{f_{\theta}}(\Nmu_{l+1/2})\right) \NSigma(lh/L) \left(\1_{D\times D} + \frac{h}{L}  J_{f_{\theta}}(\Nmu_{l+1/2})\right)^{\top}\\
&\quad\quad\quad +\frac{h}{L} \left(\1_{D\times D} + \frac{h}{2L}  J_{f_{\theta}}(\Nmu_{l+1/2})\right) \sigma_{\theta}(\Nmu_{l+1/2})\sigma_{\theta}^{\top}(\Nmu_{l+1/2}) \left(\1_{D\times D} + \frac{h}{2L}  J_{f_{\theta}}(\Nmu_{l+1/2})\right)^{\top},\\
\end{aligned}
\end{equation*}
satisfies
\begin{equation*}
\Nmu(h/L) - \Nmu_{l+1} = \mathcal{O}((h/L)^3),\quad
\NSigma(h/L)-\NSigma_{l+1} = \mathcal{O}(\norm{\sigma_{\theta}}_F^2(h/L)^3).
\end{equation*}
Applying the midpoint scheme to (\ref{eq: gauss discretization2}), we have
\begin{equation*}
\begin{aligned}
&\bar{\mu}_{l+1/2} = \Nmu(lh/L) + \frac{h}{2L} f_{\theta}(\Nmu(lh/L)), \quad
\bar{\mu}_{l+1} = \Nmu(lh/L) + \frac{h}{L} f_{\theta}(\bar{\mu}_{l+1/2});\\
&\bar{\Sigma}_{l+1/2} = \NSigma(lh/L) + \frac{h}{2L} \brac{J_{f_{\theta}}(\Nmu(lh/L)) \NSigma(lh/L) + \NSigma(lh/L) J^{\top}_{f_{\theta}}(\Nmu(lh/L)) + \sigma_{\theta}(\Nmu(lh/L))\sigma_{\theta}^{\top}(\Nmu(lh/L))}, \\
&\bar{\Sigma}_{l+1} = \NSigma(lh/L) + \frac{h}{L} \brac{J_{f_{\theta}}(\bar{\mu}_{l+1/2}) \NSigma_{l+1/2} + \NSigma_{l+1/2} J^{\top}_{f_{\theta}}(\bar{\mu}_{l+1/2}) + \sigma_{\theta}(\bar{\mu}_{l+1/2})\sigma_{\theta}^{\top}(\bar{\mu}_{l+1/2})}.
\end{aligned}
\end{equation*}
The fact that $\bar{\mu}_{l+1} = \Nmu_{l+1}$ yields the first estimates immediately. In addition, we have that
\begin{equation*}
\begin{aligned}
\bar{\Sigma}_{l+1} - \NSigma_{l+1} =& \frac{h^2}{2L^2} \brac{J_{f_{\theta}}(\bar{\mu}_{l+1/2}) J_{f_{\theta}}(\Nmu(lh/L)) \NSigma(lh/L) + \NSigma(lh/L) J^{\top}_{f_{\theta}}(\Nmu(lh/L))J^{\top}_{f_{\theta}}(\bar{\mu}_{l+1/2})} \\
&+ \frac{h^2}{2L^2} \brac{J_{f_{\theta}}(\bar{\mu}_{l+1/2})\NSigma(lh/L) J^{\top}_{f_{\theta}}(\Nmu(lh/L)) + J_{f_{\theta}}(\Nmu(lh/L)) \NSigma(lh/L)J^{\top}_{f_{\theta}}(\bar{\mu}_{l+1/2})}\\
& - \frac{h^2}{L^2}  J_{f_{\theta}}(\Nmu_{l+1/2}) \NSigma(lh/L) J^{\top}_{f_{\theta}}(\Nmu_{l+1/2}) -\frac{h^3}{4L^3} J_{f_{\theta}}(\Nmu_{l+1/2}) \sigma_{\theta}(\Nmu_{l+1/2})\sigma_{\theta}^{\top}(\Nmu_{l+1/2})   J_{f_{\theta}}(\Nmu_{l+1/2})^{\top}.
\end{aligned}
\end{equation*}
According to the facts that $\NSigma(lh/L) = \mathcal{O}(\norm{\sigma_{\theta}}_F^2h)$ and $\bar{\Sigma}_{l+1}$ is of order-2 approximation, we conclude the second estimate. Using the classical numerical analysis results \cite[Theorem 3.4, Page 160]{hairer1993solving}, the multi-step error reads
\begin{equation*}
\Nmu(h) - \Nmu_{L} = \mathcal{O}(h^3/L^2),\quad
\NSigma(h)-\NSigma_{L} = \mathcal{O}(\norm{\sigma_{\theta}}_F^2h^3/L^2).
\end{equation*}
Subsequently, we have
\begin{equation*}
\begin{aligned}
\E{\norm{X_h - X_G}_2^2} &= \E{\|\Nmu_{L} + \sqrt{\NSigma_{L}} \N(\0, \1) - \Nmu(h)- \sqrt{\NSigma(h)} \N(\0, \1)\|_2^2}\\
&=(\Nmu_{L}- \Nmu(h))^2 + (\sqrt{\NSigma_{L}}-\sqrt{\NSigma(h)})^2 = \mathcal{O}(h^6/L^4) + \mathcal{O}(\norm{\sigma_{\theta}}_F^2h^5/L^4),
\end{aligned}
\end{equation*}
where we again used the fact that $\NSigma(h)=\mathcal{O}(\norm{\sigma_{\theta}}_F^2h)$. Combining the random perturbation theory of dynamical systems, \cref{cor:gauss}, we conclude the proof.
\end{proof}

\begin{proof}[Proof of \cref{the:dis con2}]
Observing the fact that the scheme (\ref{eq:ite of den cov2}) is applying the explicit midpoint scheme to the mean and employing the modified Euler scheme to the covariance, we can obtain the numerical error immediately:
\begin{equation*}
\Nmu(h) - \Nmu_{L} = \mathcal{O}(h^3),\quad
\NSigma(h)-\NSigma_{L} = \mathcal{O}(\norm{\sigma_{\theta}}_F^2h^2).
\end{equation*}
Then the estimate in \cref{the:dis con2} is derived using the same method as outlined in the proof of \cref{the:dis con}.
\end{proof}

\begin{proof}[Proof of \cref{the:dis con3}]
The estimate in \cref{the:dis con3} is also obtained as in the proof of \cref{the:dis con}, by observing the fact that Gaussian approximation (\ref{eq:GaussApp}) derived through the Euler-Maruyama scheme is applying the one-step Euler scheme to (\ref{eq: gauss discretization2}).
\end{proof}

\bibliographystyle{abbrv}
\bibliography{ref}

\begin{thebibliography}{10}

\bibitem{ait2002maximum}
Y.~A{\"\i}t-Sahalia.
\newblock Maximum likelihood estimation of discretely sampled diffusions: a closed-form approximation approach.
\newblock {\em Econometrica}, 70(1):223--262, 2002.

\bibitem{arasaratnam2009cubature}
I.~Arasaratnam and S.~Haykin.
\newblock Cubature kalman filters.
\newblock {\em IEEE Transactions on automatic control}, 54(6):1254--1269, 2009.

\bibitem{bertalan2019learning}
T.~Bertalan, F.~Dietrich, I.~Mezi{\'c}, and I.~G. Kevrekidis.
\newblock On learning hamiltonian systems from data.
\newblock {\em Chaos: An Interdisciplinary Journal of Nonlinear Science}, 29(12), 2019.

\bibitem{blagoveshchenskii1962diffusion}
Y.~N. Blagoveshchenskii.
\newblock Diffusion processes depending on a small parameter.
\newblock {\em Theory of Probability \& Its Applications}, 7(2):130--146, 1962.

\bibitem{blagoveshchenskii1961certain}
Y.~N. Blagoveshchenskii and M.~I. Freidlin.
\newblock Certain properties of diffusion processes depending on a parameter.
\newblock In {\em Soviet Math. Dokl}, volume~2, pages 633--636, 1961.

\bibitem{bongard2007automated}
J.~Bongard and H.~Lipson.
\newblock Automated reverse engineering of nonlinear dynamical systems.
\newblock {\em Proceedings of the National Academy of Sciences}, 104(24):9943--9948, 2007.

\bibitem{brandt2002simulated}
M.~W. Brandt and P.~Santa-Clara.
\newblock Simulated likelihood estimation of diffusions with an application to exchange rate dynamics in incomplete markets.
\newblock {\em Journal of financial economics}, 63(2):161--210, 2002.

\bibitem{brunton2017chaos}
S.~L. Brunton, B.~W. Brunton, J.~L. Proctor, E.~Kaiser, and J.~N. Kutz.
\newblock Chaos as an intermittently forced linear system.
\newblock {\em Nature communications}, 8(1):1--9, 2017.

\bibitem{chekroun2011stochastic}
M.~D. Chekroun, E.~Simonnet, and M.~Ghil.
\newblock Stochastic climate dynamics: Random attractors and time-dependent invariant measures.
\newblock {\em Physica D: Nonlinear Phenomena}, 240(21):1685--1700, 2011.

\bibitem{chen2022automated}
B.~Chen, K.~Huang, S.~Raghupathi, I.~Chandratreya, Q.~Du, and H.~Lipson.
\newblock Automated discovery of fundamental variables hidden in experimental data.
\newblock {\em Nature Computational Science}, 2(7):433--442, 2022.

\bibitem{chen2023constructing}
X.~Chen, B.~W. Soh, Z.-E. Ooi, E.~Vissol-Gaudin, H.~Yu, K.~S. Novoselov, K.~Hippalgaonkar, and Q.~Li.
\newblock Constructing custom thermodynamics using deep learning.
\newblock {\em Nature Computational Science}, pages 1--20, 2023.

\bibitem{chen2021solving}
X.~Chen, L.~Yang, J.~Duan, and G.~E. Karniadakis.
\newblock Solving inverse stochastic problems from discrete particle observations using the fokker--planck equation and physics-informed neural networks.
\newblock {\em SIAM Journal on Scientific Computing}, 43(3):B811--B830, 2021.

\bibitem{chen2023learning}
Y.~Chen and D.~Xiu.
\newblock Learning stochastic dynamical system via flow map operator.
\newblock {\em arXiv preprint arXiv:2305.03874}, 2023.

\bibitem{dietrich2023learning}
F.~Dietrich, A.~Makeev, G.~Kevrekidis, N.~Evangelou, T.~Bertalan, S.~Reich, and I.~G. Kevrekidis.
\newblock Learning effective stochastic differential equations from microscopic simulations: Linking stochastic numerics to deep learning.
\newblock {\em Chaos: An Interdisciplinary Journal of Nonlinear Science}, 33(2), 2023.

\bibitem{greydanus2019hamiltonian}
S.~Greydanus, M.~Dzamba, and J.~Yosinski.
\newblock {H}amiltonian neural networks.
\newblock In {\em 33rd Conference on Neural Information Processing Systems (NeurIPS 2019)}, pages 15353--15363, 2019.

\bibitem{gu2023stationary}
Y.~Gu, J.~Harlim, S.~Liang, and H.~Yang.
\newblock Stationary density estimation of it{\^o} diffusions using deep learning.
\newblock {\em SIAM Journal on Numerical Analysis}, 61(1):45--82, 2023.

\bibitem{hairer1993solving}
E.~Hairer, S.~P. Nørsett, and G.~Wanner.
\newblock {\em Solving ordinary differential equations I: Nonstiff Problems}.
\newblock Springer Berlin, Heidelberg, 1993.

\bibitem{hamzi2023note}
B.~Hamzi, H.~Owhadi, and L.~Paillet.
\newblock A note on microlocal kernel design for some slow-fast stochastic differential equations with critical transitions and application to eeg signals.
\newblock {\em PHYSICA A-STATISTICAL MECHANICS AND ITS APPLICATIONS}, 616, 2023.

\bibitem{heerboth2015emt}
S.~Heerboth, G.~Housman, M.~Leary, M.~Longacre, S.~Byler, K.~Lapinska, A.~Willbanks, and S.~Sarkar.
\newblock Emt and tumor metastasis.
\newblock {\em Clinical and translational medicine}, 4(1):1--13, 2015.

\bibitem{iacus2008simulation}
S.~M. Iacus.
\newblock {\em Simulation and inference for stochastic differential equations: with R examples}, volume 486.
\newblock Springer, 2008.

\bibitem{jensen2002transition}
B.~Jensen and R.~Poulsen.
\newblock Transition densities of diffusion processes: numerical comparison of approximation techniques.
\newblock {\em Journal of Derivatives}, 9(4):18, 2002.

\bibitem{kessler1997estimation}
M.~Kessler.
\newblock Estimation of an ergodic diffusion from discrete observations.
\newblock {\em Scandinavian Journal of Statistics}, 24(2):211--229, 1997.

\bibitem{kinga2015method}
D.~Kinga, J.~B. Adam, et~al.
\newblock A method for stochastic optimization.
\newblock In {\em International conference on learning representations (ICLR)}, volume~5, page~6. San Diego, California;, 2015.

\bibitem{kingma2014adam}
D.~P. Kingma and J.~Ba.
\newblock Adam: {A} method for stochastic optimization.
\newblock In {\em 3rd International Conference on Learning Representations}, 2015.

\bibitem{kushner1967approximations}
H.~Kushner.
\newblock Approximations to optimal nonlinear filters.
\newblock {\em IEEE Transactions on Automatic Control}, 12(5):546--556, 1967.

\bibitem{li2018landscape}
C.~Li and G.~Balazsi.
\newblock A landscape view on the interplay between emt and cancer metastasis.
\newblock {\em NPJ systems biology and applications}, 4(1):34, 2018.

\bibitem{li2021numerical}
L.~Li, J.~Lu, J.~C. Mattingly, and L.~Wang.
\newblock Numerical methods for stochastic differential equations based on gaussian mixture.
\newblock {\em Communications in Mathematical Sciences}, 19(6):1549--1577, 2021.

\bibitem{li2020scalable}
X.~Li, T.~L. Wong, R.~T.~Q. Chen, and D.~Duvenaud.
\newblock Scalable gradients for stochastic differential equations.
\newblock In {\em The 23rd International Conference on Artificial Intelligence and Statistics, {AISTATS} 2020}, volume 108 of {\em Proceedings of Machine Learning Research}, pages 3870--3882. {PMLR}, 2020.

\bibitem{lin2023computing}
B.~Lin, Q.~Li, and W.~Ren.
\newblock Computing high-dimensional invariant distributions from noisy data.
\newblock {\em Journal of Computational Physics}, 474:111783, 2023.

\bibitem{look2019differential}
A.~Look and M.~Kandemir.
\newblock Differential bayesian neural nets.
\newblock In {\em Proc. Adv. Neural Informat. Process. Syst. Workshop Bayesian Deep Learn., 2019.}, 2019.

\bibitem{look2022deterministic}
A.~Look, M.~Kandemir, B.~Rakitsch, and J.~Peters.
\newblock A deterministic approximation to neural sdes.
\newblock {\em IEEE Transactions on Pattern Analysis and Machine Intelligence}, 45(4):4023--4037, 2023.

\bibitem{lorenz1963deterministic}
E.~N. Lorenz.
\newblock Deterministic nonperiodic flow.
\newblock {\em Journal of atmospheric sciences}, 20(2):130--141, 1963.

\bibitem{makeev2002coarse}
A.~G. Makeev, D.~Maroudas, and I.~G. Kevrekidis.
\newblock “coarse” stability and bifurcation analysis using stochastic simulators: Kinetic monte carlo examples.
\newblock {\em The Journal of chemical physics}, 116(23):10083--10091, 2002.

\bibitem{mcnamee1967construction}
J.~McNamee and F.~Stenger.
\newblock Construction of fully symmetric numerical integration formulas of fully symmetric numerical integration formulas.
\newblock {\em Numerische Mathematik}, 10:327--344, 1967.

\bibitem{pavliotis2016stochastic}
G.~A. Pavliotis.
\newblock {\em "Markov Processes and the Chapman–Kolmogorov Equation". Stochastic processes and applications}.
\newblock Springer, 2016.

\bibitem{pedersen1995new}
A.~R. Pedersen.
\newblock A new approach to maximum likelihood estimation for stochastic differential equations based on discrete observations.
\newblock {\em Scandinavian journal of statistics}, pages 55--71, 1995.

\bibitem{prakasa1999statistical}
B.~Prakasa~Rao.
\newblock {\em Statistical inference for diffusion type processes}.
\newblock Oxford University Press, New York, 1999.

\bibitem{rudy2017data}
S.~H. Rudy, S.~L. Brunton, J.~L. Proctor, and J.~N. Kutz.
\newblock Data-driven discovery of partial differential equations.
\newblock {\em Science Advances}, 3(4):e1602614, 2017.

\bibitem{sarkka2013gaussian}
S.~S{\"a}rkk{\"a} and J.~Sarmavuori.
\newblock Gaussian filtering and smoothing for continuous-discrete dynamic systems.
\newblock {\em Signal Processing}, 93(2):500--510, 2013.

\bibitem{sarkka2023bayesian}
S.~S{\"a}rkk{\"a} and L.~Svensson.
\newblock {\em Bayesian filtering and smoothing}, volume~17.
\newblock Cambridge university press, 2023.

\bibitem{schmidt2009distilling}
M.~Schmidt and H.~Lipson.
\newblock Distilling free-form natural laws from experimental data.
\newblock {\em Science}, 324(5923):81--85, 2009.

\bibitem{solin2021scalable}
A.~Solin, E.~Tamir, and P.~Verma.
\newblock Scalable inference in sdes by direct matching of the fokker--planck--kolmogorov equation.
\newblock {\em Advances in Neural Information Processing Systems}, 34:417--429, 2021.

\bibitem{song2021score}
Y.~Song, J.~Sohl{-}Dickstein, D.~P. Kingma, A.~Kumar, S.~Ermon, and B.~Poole.
\newblock Score-based generative modeling through stochastic differential equations.
\newblock In {\em 9th International Conference on Learning Representations, {ICLR} 2021, Virtual Event, Austria, May 3-7, 2021}. OpenReview.net, 2021.

\bibitem{sparrow2012lorenz}
C.~Sparrow.
\newblock {\em The Lorenz equations: bifurcations, chaos, and strange attractors}, volume~41.
\newblock Springer Science \& Business Media, 2012.

\bibitem{tzen2019neural}
B.~Tzen and M.~Raginsky.
\newblock Neural stochastic differential equations: Deep latent gaussian models in the diffusion limit.
\newblock {\em arXiv preprint arXiv:1905.09883}, 2019.

\bibitem{williams1989learning}
R.~J. Williams and D.~Zipser.
\newblock A learning algorithm for continually running fully recurrent neural networks.
\newblock {\em Neural computation}, 1(2):270--280, 1989.

\bibitem{xu2023modeling}
Z.~Xu, Y.~Chen, Q.~Chen, and D.~Xiu.
\newblock Modeling unknown stochastic dynamical system via autoencoder.
\newblock {\em arXiv preprint arXiv:2312.10001}, 2023.

\bibitem{yu2021onsagernet}
H.~Yu, X.~Tian, E.~Weinan, and Q.~Li.
\newblock Onsagernet: Learning stable and interpretable dynamics using a generalized onsager principle.
\newblock {\em Physical Review Fluids}, 6(11):114402, 2021.

\bibitem{zhang2022gfinns}
Z.~Zhang, Y.~Shin, and G.~Em~Karniadakis.
\newblock Gfinns: Generic formalism informed neural networks for deterministic and stochastic dynamical systems.
\newblock {\em Philosophical Transactions of the Royal Society A}, 380(2229):20210207, 2022.

\end{thebibliography}
\end{document}